\documentclass[conference]{IEEEtran}
\usepackage{amssymb,amsmath,amsthm}
\usepackage{cases}
\usepackage{graphicx}
\usepackage{stmaryrd}
\SetSymbolFont{stmry}{bold}{U}{stmry}{m}{n}
\usepackage{multirow,url}
\usepackage{color,cite}
\usepackage{cite}
\usepackage{tikz,pgf}
\usetikzlibrary{fadings,shapes,arrows,shadows}
\usepackage{makecell}
\usepackage{array}
\usepackage{overpic} 
\usepackage{subfigure}
\usepackage{url}
\usepackage{float}
\usepackage{booktabs} 
\usepackage{pifont}
\usepackage{bbm}
\usepackage{authblk}
\usepackage[linesnumbered,ruled,vlined]{algorithm2e}
\usepackage[normalem]{ulem}
\newcommand{\zxxy}[1]{{\color{black}#1}} 

\newtheorem{theorem}{Theorem}

\theoremstyle{plain}

\IEEEoverridecommandlockouts

\begin{document}

\title{A Novel Hybrid Inference Scheme for Diffusion-Based AIGC Services in MEC Networks}

\author{Xinyi Zhuang,
			Jiaqi Wu,
			Yuan Luo, \emph{Senior Member, IEEE},
			Ming Tang,
			Huaizhe Liu,
			Hongjia Wu, \\
			Lin Gao, \textit{Senior Member, IEEE},
			and Qinyu Zhang, \textit{Senior Member, IEEE}
\thanks{Part of the results have been accepted by IEEE ICC 2025 \cite{ZXY-ICC}.}
\thanks{This work was supported in part by the Natural Science Foundation of Guangdong Province (Grant No. 2024A1515010178), the Shenzhen Science and Technology Program (Grant Nos. GXWD20231129103946001, KJZD20240903095402004, JCYJ20230807114300001, and ZDYJ20251211140205007), and the National Natural Science Foundation of China (Grant Nos. 62672225 and 62472367).
This work was also supported in part by ZTE Industry-University-Institute Cooperation Funds under Grant No. IA20260728006-V1.
This work was also supported in part by the Focused Project (FPF10120250010) and the Cultivation-oriented Seed Project (SPF10120250004) of the Shenzhen Loop Area Institute.}
\thanks{X. Zhuang, H. Liu, L. Gao, and Q. Zhang are with the School of Information
Science and Technology, and the Guangdong Provincial Key Laboratory of
Aerospace Communication and Networking Technology, Harbin Institute of
Technology, Shenzhen, China.
L. Gao is also with the Shenzhen Loop Area Institute, Shenzhen, China.
J. Wu is with the School of Computer Science, Guangdong University of Finance, Guangzhou, China.
Y. Luo is with the School of Science and Engineering, The Chinese University of Hong Kong, Shenzhen, China.
M. Tang is with the Department of Computer Science and Engineering, Southern University of Science and Technology, Shenzhen, China.
H. Wu is with the Department of Mathematics and Information Technology, The Education University of Hong Kong, Hong Kong, China.
Email: gaol@hit.edu.cn. \textit{(Corresponding Author: Lin Gao)}}}

\maketitle

\addtolength{\abovedisplayskip}{-1mm}
\addtolength{\belowdisplayskip}{-1mm}

\begin{abstract}
	AI-Generated Content (AIGC) powered by Generative Diffusion Models (GDMs) is transforming content creation and driving significant advancements across various fields.
	With the advancement of Mobile Edge Computing (MEC) and model compression techniques, GDMs are increasingly being deployed on Edge Servers (ESs) and User Equipment (UEs), offering more adaptive and flexible AIGC services.
	In the MEC scenario, however, designing an efficient inference
	scheme remains a significant challenge, due to the diverse
	intentions of users, as well as the resource limitations of ESs and UEs.
	In this work, we propose a novel \emph{Hybrid Inference Scheme (HIS)} for MEC-enabled AIGC services, consisting of two phases:
	(i) a \emph{common inference phase} that  generates shared intermediate results for users with similar intentions using  public prompts derived from their provided personal prompts, and
	(ii) a \emph{personalized inference phase} that generates final results (contents) for each individual user based on their personal prompts.
	To quantitatively measure the service quality of the HIS, we derive a novel Hybrid Inference Quality Metric (HIQM) through real-world data, which measures the semantic alignment of generated content across varying semantic similarities and common inference steps.
	Building upon this metric, we formulate a joint optimization problem that integrates user clustering, model inferencing, and resource allocation, aiming to  optimize service quality, delay, and energy consumption jointly.
	We further propose a semantic-based user clustering approach, and an integrated deep reinforcement learning and optimization approach for model inferencing and resource allocation.
	Simulation results demonstrate that our  approach  significantly outperforms existing benchmarks, achieving performance improvements ranging from 36.8\% to 94.3\%.
\end{abstract}

\begin{IEEEkeywords}
	AI-generated content, generative diffusion model, mobile edge computing, deep reinforcement learning\vspace{4mm}
\end{IEEEkeywords}

\IEEEpeerreviewmaketitle

\section{Introduction}\label{sec:Introduction}

\subsection{Background and Motivations}
AI-Generated Content (AIGC) has recently attracted significant attention for its transformative impact on controllable content generation across multiple domains.
A key driver behind this revolution is the development of large-scale AI models that exhibit exceptional capabilities in synthesizing high-quality images, videos, and 3D representations. 
Among these models, Generative Diffusion Models (GDMs) have emerged as one of the most influential classes. In particular, GDMs play a pivotal role in a wide range of interactive applications,  such as point cloud reconstruction, autonomous driving, robotics, and artistic expression \cite{AIGC-Survey}.

The success of GDMs lies in their unique inference mechanism that endows them with remarkable generative capabilities.
Specifically, GDMs employ a series of step-wise and reversible denoising processes (called \textit{inference steps}) that progressively transform a simple
noise distribution into structured data distributions.
This process---effectively a discrete solution to a Stochastic Differential Equation (SDE) \cite{Song-Yang-Score}---enables fine-grained control over content generation.
This groundbreaking capability has led to many successful commercial models such as Stable Diffusion and Sora, demonstrating the great potential of GDMs in both research and industry.
As a result, the continuous evolution of GDMs is expected to drive further innovations across various application domains, particularly in wireless networks with the explosion of AI  applications.~~~~~~~~~~~~

With the growing deployment of GDMs in wireless networks, new challenges regarding stability, robustness, and efficiency are emerging. Specifically, GDMs are commonly deployed on \emph{cloud servers} to deliver large-scale AIGC services. 
However, accessing AIGC services from remote cloud servers poses several challenges, such as server vulnerability, service security, and latency issues, which limit their application potential in future wireless networks.
To overcome these limitations, \textit{Mobile Edge Computing (MEC)} \cite{add-mec01,add-mec02,add-mec03,add-mec04} has emerged as a crucial enabling technology for AIGC services.
By offloading computation to Edge Servers (ESs) and User Equipment (UEs) that are close to end users, MEC can effectively reduce service delay and enhance inference efficiency \cite{add-11,add-12,add-13,add-14}.
While model compression techniques have facilitated the deployment of GDMs on   ESs and UEs, it remains a significant challenge to design efficient inference schemes for GDMs in resource-constrained MEC environments, especially given the heterogeneity of user intentions.

Prior works \cite{Pruning-DDSM, eDiff-I, OMS-DPM, Hybrid-SD, XZH-Sharing} have proposed to improve the inference efficiency through optimizing model inferencing and task offloading.
However, these approaches mainly focus on \emph{independent inference} schemes, where the inference processes of different users are optimized separately and independently, without considering and leveraging the similarities among user intentions to enable collaborative inference. 
Given the inherent nature of GDMs (i.e., evolving from simple noise toward diverse outputs), users with similar intentions can potentially collaborate (e.g., sharing early inference steps) to reduce redundant computation and enhance inference efficiency.
However, current \emph{collaborative inference} schemes \cite{DHY-Sharing1, DHY-Sharing2, XGC-Sharing} often resort to simplistic designs, such as randomly selecting one user to perform inference and share the intermediate result with other users.
Obviously, these approaches do not fully explore and exploit the similarities among user intentions.
This motivates us to explore a more efficient collaborative inference scheme that leverages the intention similarities among users, thus leading to the following challenge:

\vspace{0.25\baselineskip}

{\bf{Challenge 1.}} \emph{How to improve the inference efficiency by leveraging the diverse similarities among user intentions to fully exploit the potential of collaborative inference?}
\vspace{0.25\baselineskip}

Furthermore, due to resource constraints on ESs and UEs, the joint optimization of model inferencing and resource allocation is essential for enhancing inference efficiency.
Some works (e.g., \cite{LHZ-WiOpt, ZXY-INFOCOM}) focused on optimizing the number of inference steps to balance service quality, delay, and energy consumption, while others (e.g., \cite{DNN-Partitioning, CPU-Frequency-Allocation}) explored GPU or CPU-based frequency scaling for fine-grained resource control. 
However, these approaches are based on independent inference schemes, overlooking the potential benefits of collaborative inference among different users. 

In addition, in the context of collaborative inference,   similarities among user intentions play a critical role in the results.
Specifically, for users with similar intentions, collaborative inference can  significantly improve the inference efficiency, while preserving the quality of   generated contents.
For users with diverse intentions, however, collaborative inference may lead to severe quality degradation.
Therefore, it is crucial to design an effective user clustering algorithm that groups users with similar intentions together.
This motivates us to optimize user clustering together with model inferencing and resource allocation, thus giving rise to  the following challenge:

\vspace{0.25\baselineskip}
{\bf{Challenge 2.}} \emph{How to further improve the inference efficiency by jointly optimizing user clustering, model inferencing, and resource allocation for collaborative inference?}
\vspace{0.25\baselineskip}

To address these challenges, we develop an efficient collaborative inference scheme that leverages the similarities among user intentions, and propose effective algorithms for user clustering, model inferencing, and resource allocation.

\subsection{Solution and Contributions}
\begin{figure}[t]
	\centering
	\includegraphics[width=3.4in]{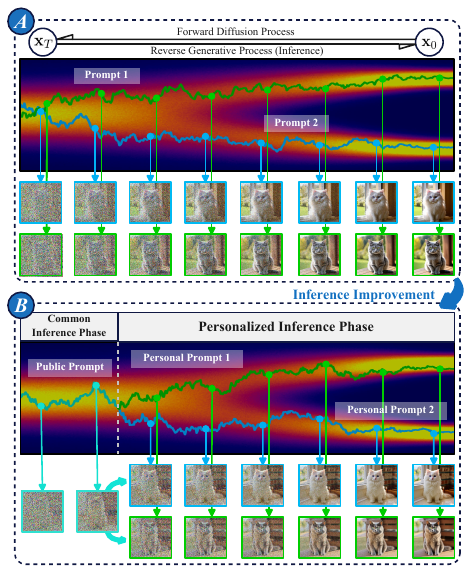}
	\vspace{-4mm}
	\caption{A synthetic overview of the traditional independent inference scheme (\textbf{Part A}) and our proposed HIS (\textbf{Part B}).}
	\label{fig:1-DM-HIS}
	\vspace{0mm}
\end{figure}

\textbf{To address challenge 1}, we propose a novel \emph{Hybrid Inference Scheme (HIS)} to leverage intention similarities among users to improve inference efficiency. The HIS consists of two phases:
(i) a \emph{common inference phase}, where common intermediate results are generated for users with similar intentions using public prompts derived from their individual personal prompts, effectively capturing the general characteristics of these prompts; 
(ii) a \emph{personalized inference phase}, where final results (contents) are generated for users based on their specific personal prompt.
To further enhance efficiency, we develop a Semantic Intensity Modulator (SIM) for the common inference phase to improve the HIS performance.
For clarity, we compare  the traditional independent inference scheme and our proposed HIS in \mbox{Fig. \ref{fig:1-DM-HIS}}. 
In the traditional scheme (Part A), inference processes operate independently using personal prompts.
In contrast, in our proposed HIS (Part B), a common inference process first generates the common intermediate results using public prompts, and then two personalized inference processes generate the final results using their personal prompts.  

To evaluate the service quality of the HIS, we propose a new \emph{Hybrid Inference Quality Metric (HIQM)} based on extensive real-world experiments. 
This metric is designed to quantify the semantic alignment between the generated content and the corresponding prompt, taking into account different levels of semantic similarity and the number of common inference steps. 
The introduction of HIQM provides a precise, quantitative measure to assess the effectiveness of HIS across varying scenarios and user intentions.


\textbf{To address challenge 2}, we focus on optimizing the inference efficiency in a resource-constrained MEC network consisting of one ES and multiple UEs, where both the ES and UEs are equipped with GDMs to collaboratively deliver AIGC services.
In such a scenario, our goal is to jointly optimize user clustering, model inferencing, and resource allocation, aiming to enhance service quality, reduce service delay, and minimize energy consumption.
For user clustering, we propose a semantic-based user clustering approach by refining a Hierarchical Agglomerative Clustering (HAC) algorithm, which groups semantically similar users into the same cluster. 
For model inferencing and resource allocation, we propose a two-tier solution that integrates both learning and optimization techniques. 
In the first tier, we design a Deep Reinforcement Learning (DRL)-based Independent Proximal Policy Optimization (IPPO) approach to optimize the common and personalized inference steps.
In the second tier, we derive closed-form solutions for the optimal resource allocation based on convex optimization theory.
In summary, the key contributions of this work are summarized as follows:

\begin{itemize}
	\vspace{-0mm}
	\item {\textbf{\textit{Novel   Inference Scheme:}}}
	We propose a novel hybrid inference scheme, HIS, which exploits user intention similarities to improve inference efficiency.
	We also propose a new performance metric, HIQM, to evaluate the alignment of generated contents with prompts.
	\item {\textbf{\textit{Optimization of User Clustering:}}}
	\zxxy{We propose a semantic-based user clustering approach, which improves overall alignment between public and personal prompts in each cluster, thus enhancing inference efficiency through enabling more common inference steps.}
	\item {\textbf{\textit{Joint Optimization of Model Inferencing and Resource Allocation:}}}
	We propose a two-tier solution that combines both learning and optimization techniques for the joint optimization of model inferencing and resource allocation.
	The approach exhibits exceptional flexibility and scalability, making it well-suited for   real-world scenarios.
	\item {\textbf{\textit{Performance Evaluation:}}}
	Simulation results demonstrate that our proposed scheme significantly outperforms existing benchmarks, with performance improvements ranging from 36.8\% to 94.3\%, highlighting its superiority in terms of efficiency and robustness.
\end{itemize}

\section{Related Work}\label{sec:RelatedWork}

\subsection{Model Acceleration}
GDMs have achieved state-of-the-art performance across various domains. 
Current best-performing GDMs mainly rely on U-Net or transformer-based architectures, which involve iterative evaluations of large-scale neural networks, resulting in significant computational overhead.

Low-resource inference for efficient GDMs has recently attracted substantial interest.
One major direction focuses on reducing the total number of inference steps.
For instance, DDIM \cite{DDIM} and DPM solver \cite{DPMSolver} have introduced first-order and high-order solvers to accelerate the inference process.
Parallel efforts explore architectural compression: knowledge distillation \cite{Knowledge-Distillation} trains lightweight student models to mimic pre-trained teachers, quantization techniques \cite{Quantization} reduce numerical precision of weights or activations, and pruning \cite{Pruning-DDSM} eliminates redundant network parameters.
Additionally, the latent diffusion framework \cite{Latent-Space} significantly reduces computational costs by operating in a compressed latent feature space rather than pixel domain.
However, most existing approaches necessitate computationally intensive retraining or fine-tuning to sustain satisfactory performance, which imposes practical limitations in resource-constrained scenarios.

In our work, we tackle the computational challenge from a unique angle.
We aim to curtail the overall computation of GDMs by sharing a part of the inference process.
It is crucial to emphasize that our proposed scheme is \textit{orthogonal} to the aforementioned methods and can seamlessly integrate with them without inducing extra cost.

\subsection{Collaborative Inference}\label{sec:RelatedWork-CollaborativeInference}
In the context of well-trained GDMs, some researchers have explored model scheduling to alleviate the computational overhead.
As one of the pioneering works, NVIDIA Corp. \cite{eDiff-I} introduced the eDiff-I framework, which decomposes the inference process into multiple stages, each handled by a distinct model to process hierarchical information.
Following this, Yang \textit{et al.} \cite{Pruning-DDSM} and Liu \textit{et al.} \cite{OMS-DPM} further advanced this approach by scheduling varying-scale GDMs across different inference steps to accelerate inference.
However, these works focus on pre-defined computational graphs and cannot adapt to dynamic network conditions.
In contrast, ByteDance Inc. \cite{Hybrid-SD} proposed the edge-cloud framework, which splits the entire inference process into two phases: the first is handled in the cloud, while the second is executed at the edge.
Yang \textit{et al.} \cite{XZH-Sharing} further optimized the split point in an end-edge framework by considering latency constraints and computational resources.
Nonetheless,  the above works  focus on independent inference, where   inference processes of different users are independent of each other.

By leveraging user intentions, Du \textit{et al.} \cite{DHY-Sharing1} proposed the collaborative inference scheme in which multiple users share the same inference process during the initial phase to reduce redundant computation.
Their subsequent work \cite{DHY-Sharing2} introduced a two-phase architecture where the edge server first performs inference by leveraging a randomly selected prompt, followed by on-device refinement for personalization.
Xie \textit{et al.} \cite{XGC-Sharing} further extended this architecture to vehicular networks by introducing an entity-based clustering strategy.
However, these schemes oversimplify the diversity of user intentions and simply assume a homogeneous intermediate result for personalization.
This stringent setting significantly shortens the shared inference process, ultimately constraining the potential efficiency for GDM inference.

In our study, we \textit{re-design} the collaborative inference scheme by considering heterogeneous user intentions.
We further develop an advanced module for the common inference phase.
Moreover, we investigate the user clustering problem, together with model inferencing and resource allocation, to improve the inference efficiency of the proposed HIS.
It is worth noting that the HIS is orthogonal to parameter-efficient personalization methods such as LoRA-based adapters \cite{LoRA}: instead of adapting model weights for each user, it is training-free and realizes personalization entirely at the inference level, and can thus be readily combined with them.

\subsection{Model Inferencing and Resource Allocation}
For MEC-enabled AIGC services, existing works typically focused on optimizing either model inferencing or resource allocation separately.
Regarding model inferencing, several works (e.g., \cite{DNN-Partitioning}) primarily addressed the deep neural network partitioning problem.
By leveraging the new features of GDMs, Liu \textit{et al.} \cite{LHZ-WiOpt} and Zhuang \textit{et al.} \cite{ZXY-INFOCOM} proposed optimizations that reduce the total number of inference steps to enhance network utility.
However, these approaches mainly concentrate on vanilla GDMs and often overlook the impact of user intention heterogeneity.

As for resource allocation, some works (e.g., \cite{D2SAC}) have explored resource management through indirect service scheduling mechanisms.
While these approaches demonstrate some initial effectiveness, there remains significant potential for further optimization.
For example, Liu \textit{et al.} \cite{DNN-Partitioning} and Li \textit{et al.} \cite{CPU-Frequency-Allocation} have proposed fine-grained resource allocation techniques through computing frequency assignment.
However, these approaches are limited by the computational intensity of most GDMs, which require GPU-accelerated inference, rendering CPU-centric frequency scaling approaches less applicable.

In our study, we develop the \textit{joint optimization} framework for model inferencing and resource allocation, with the goal of achieving an optimal trade-off among service quality, delay, and energy consumption.

\section{Hybrid Inference Scheme} \label{sec:HIS}

\subsection{Content Generation}\label{Sec:Content-Generation}
Without loss of generality, GDMs synthesize images by learning to estimate a score function, which is defined as the gradient of the log-density of image distribution \cite{Song-Yang-Score}.
For clarity, we can interpret GDMs through the framework of SDE.
From this perspective, GDMs initiate a forward diffusion process that gradually corrupts an original image distribution $p({\bf{x}}_0)$ by adding Gaussian noise over time.
This progressive corruption eventually transforms the data into a noisy distribution $p({\bf{x}}_T)$.
Mathematically, this forward diffusion process can be expressed as:
\begin{equation}\textstyle
	d{\bf{x}}_t = {\bf{f}}({\bf{x}}_t,t) dt + g(t) d{\bf{w}}_t,
\end{equation}
where ${\bf{f}}({\bf{x}}_t,t)$ and $g(t)$ denote drift and diffusion coefficients, respectively, and $d{\bf{w}}_t$ represents the Wiener process \cite{Song-Yang-Score}.

As shown in Fig. \ref{fig:1-DM-HIS} (Part A), GDMs generate images by \emph{reversing} the forward process, which starts from a Gaussian prior and progressively refines the distribution through the estimated score function.
This reverse generative process, i.e., the inference process, can be mathematically expressed as:
\begin{equation} \label{eq:reverse-sde}\textstyle
	d{\bf{x}}_t = \left[ {\bf{f}}({\bf{x}}_t,t) - g(t)^2 \nabla_{{\bf{x}}_t} \log p({\bf{x}}_t | {\bf{p}}) \right] dt + g(t) d{\bf{\bar{w}}}_t,
\end{equation}
where $\bf{p}$ denotes the prompt embeddings, $d{\bf{\bar{w}}}_t$ denotes the Wiener process, and $\nabla_{{\bf{x}}_t} \log p({\bf{x}}_t | {\bf{p}})$ represents the conditional score function approximated by neural networks.

To enhance conditional generation performance, most GDMs adopt the classifier-free guidance \cite{CFG} technique.
This leads to a reformulated score function:
\begin{equation}\label{eq:guidance-scale}\textstyle
	\begin{aligned}
		\nabla_{{\bf{x}}_t} \log p({\bf{x}}_t | {\bf{p}}) := &\, w \nabla_{{\bf{x}}_t} \log p({\bf{x}}_t | {\bf{p}}) \\
		& +(1-w) \nabla_{{\bf{x}}_t} \log p({\bf{x}}_t),
	\end{aligned}
\end{equation}
where $w$ denotes the guidance scale.
It significantly improves generation quality by guiding the distribution towards a higher likelihood of the condition ${\bf{p}}$.
In practice, the inference process is implemented by discretizing the reverse SDE into finite steps.
At each step, the model evaluates both conditional and unconditional scores, then combines them using the guidance scale to iteratively refine the distribution.
This iterative guidance ensures that the generated output is not only coherent but also aligned with the semantic intent of the prompt.

\subsection{Hybrid Inference Scheme (HIS)}
As discussed in Section \ref{Sec:Content-Generation}, traditional GDM inference treats each prompt independently.
For each individual prompt, GDM initializes sampling from a Gaussian distribution and then employs a score function to guide the distribution toward a prompt-specific target.
This approach may result in redundant computation, particularly when processing multiple prompts that exhibit semantic similarities \cite{DHY-Sharing1}.
In such cases, these semantic similarities can be leveraged at the early phase of inference to generate a common intermediate result, which can then be further refined individually.
Clearly, identifying and exploiting these semantic similarities can substantially reduce computational cost.
Motivated by this,  we propose the HIS\footnote{Code is available at: \url{https://github.com/iimxinyi/HIS}}  shown in Fig. \ref{fig:1-DM-HIS} (Part B), a novel scheme that decomposes the inference process into two   phases:

\begin{itemize}
	\item \emph{Common Inference Phase:}
	A public prompt ${\bf{p}}^{\text{pub}}$ is first constructed to capture the general semantic characteristics common across multiple user-provided prompts.
	Then, the GDM utilizes the score function defined in Eq. \eqref{eq:guidance-scale} with $w := w_1$ and ${\bf{p}} := {\bf{p}}^{\text{pub}}$ to generate a common intermediate result, where $w_1$ is the guidance scale for the common inference phase.
	
	\item \emph{Personalized Inference Phase:}
	The GDM injects the common intermediate result along with each personal prompt\footnote{We denote the user-provided prompt as the personal prompt to distinguish it from the public prompt.} ${\bf{p}}^{\text{per}}$ and continues inference by using the score function defined in Eq. \eqref{eq:guidance-scale} with $w := w_2$ and ${\bf{p}} := {\bf{p}}^{\text{per}}$, where $w_2$ is the guidance scale for the personalized inference phase.
	This process generates the final output that matches the specific intent of each personal prompt.
\end{itemize}

\subsection{Semantic Intensity Modulator (SIM)}
\begin{figure}[t]
	\centering
	\includegraphics[width=3in]{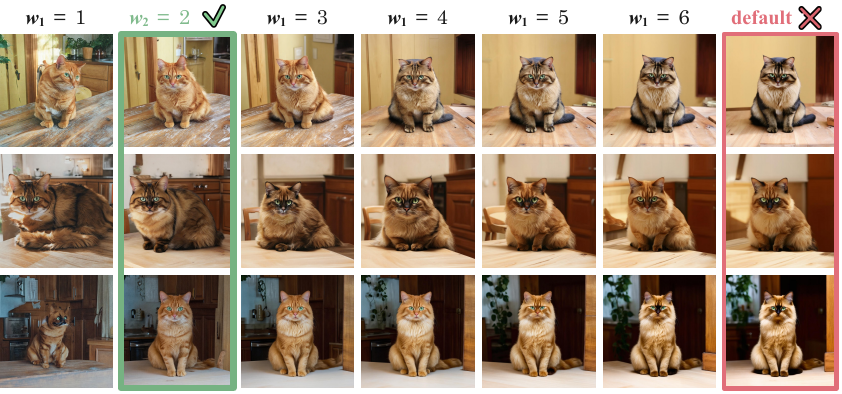}
	\vspace{-3mm}
	\caption{Generated images under different guidance scale settings. All displayed images are based on the same public prompt (``\textit{A graceful cat sitting in a warm and story-rich environment, highlighting its silky fur.}") and the same personal prompt (``\textit{A brown cat with green eyes sits calmly on a rustic wooden table in a sunlit kitchen.}")}
	\label{fig:2-SIM-Images}
	\vspace{-5mm}
\end{figure}

In the common inference phase, a key challenge is to ensure that the common intermediate result remains sufficiently compatible with a wide range of personal prompts.
As shown in the rightmost column of Fig. \ref{fig:2-SIM-Images}, the default configuration (i.e., $w_1=7.0$), which is widely used in prior works such as \cite{XGC-Sharing, Hybrid-SD, XZH-Sharing}, often yields poor performance.
The underlying reason is that the public prompt primarily offers high-level semantic information, while lacking explicit details about subjects or background elements.
As a result, during inference, GDMs often compensate for this ambiguity by autonomously ``filling in'' unspecified details.
These unintended additions in the common intermediate result introduce noise that complicates the subsequent personalized inference phase, which must first remove extraneous elements before accurately synthesizing the specified details.

\begin{figure}[t]
	\centering
	\hspace{-5mm}
	\includegraphics[width=2.8in]{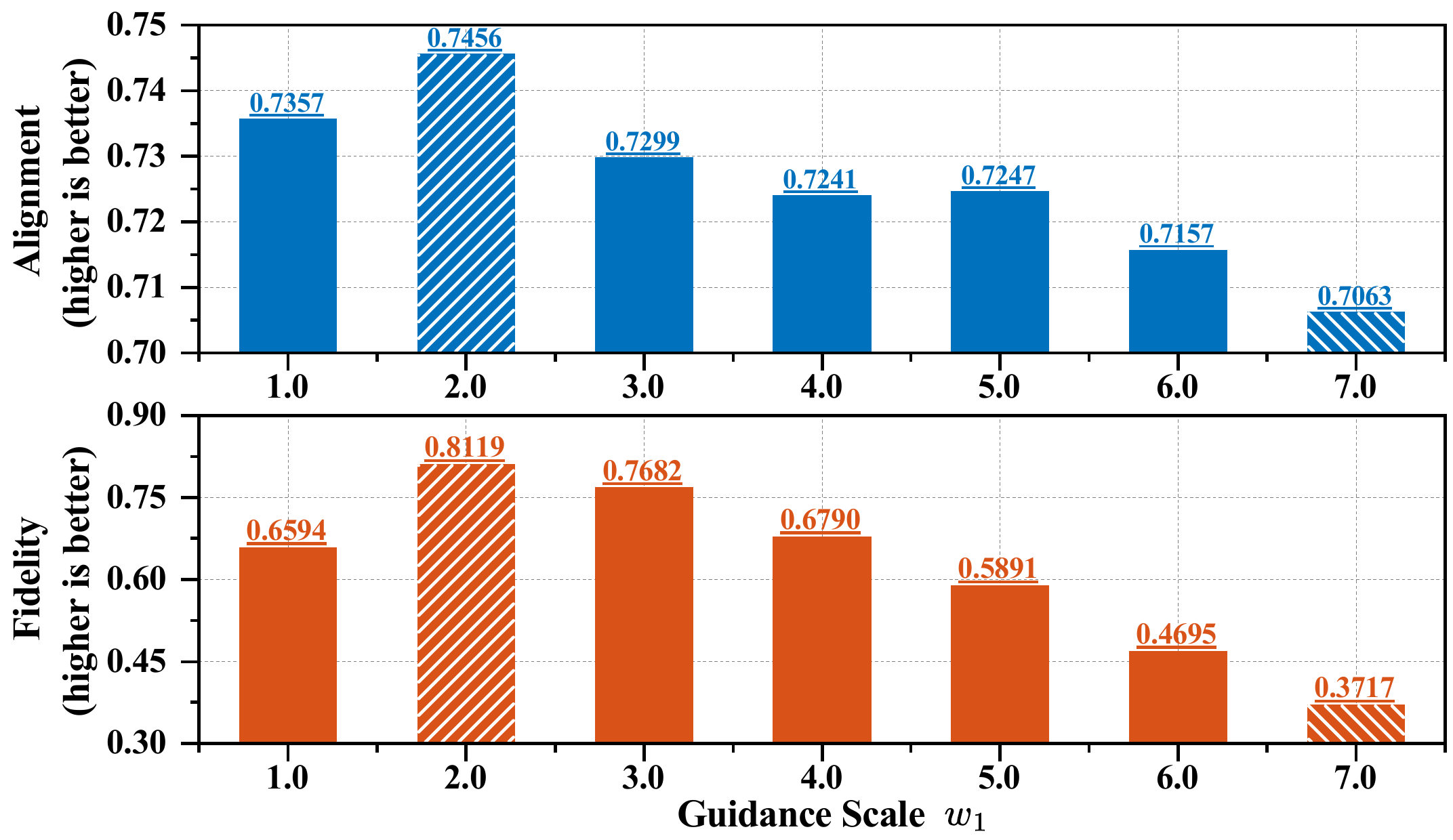}
	\vspace{-3mm}
	\caption{The alignment (\textbf{Top}) and fidelity (\textbf{Bottom}) scores under different guidance scale settings.}
	\label{fig:3-Generated-Examples}
	\vspace{-0mm}
\end{figure}

To address this challenge, we introduce the SIM\footnote{Note that the SIM is training-free and requires no modification to the underlying structure of GDMs.}, a mechanism designed to control the \emph{semantic intensity} embedded in the common intermediate result.
The key idea behind the SIM is to regulate the guidance scale $w_1$ in the common inference phase.
On one hand, as shown in Eq. \eqref{eq:guidance-scale},  lowering the guidance scale increases the randomness and naturalness in the intermediate distribution, effectively shifting it away from sharp peaks toward more central and generalized regions \cite{CFG-Fidelity}.
This transformation is particularly beneficial for subsequent personalized inference, as transitioning from a central position in the distribution space is much easier than navigating between well-separated peaks.
On the other hand, if the guidance scale is too small, the resulting intermediate distribution may diverge too far from the intended semantics.
This excessive randomness can obscure the key features of the public prompt, ultimately degrading the fidelity of the  generated images. Therefore, the SIM plays a crucial balancing role in ensuring semantic flexibility while preserving the integrity of the general intent.

To empirically investigate the above trade-off, we conduct extensive experiments using a diverse set of real-world prompts, which are drawn from both LLM-generated prompts \cite{Llama} and real user prompts in the DiffusionDB dataset \cite{DiffusionDB}.
Our goal is to quantify two critical aspects, i.e., \emph{alignment} and \emph{fidelity}, under different guidance scales.
Specifically, the alignment reflects the semantic correspondence between the final generated content and its corresponding personal prompt.
We evaluate it as the normalized average of the scores obtained by the Contrastive Language-Image Pre-Training (CLIP) model \cite{CLIP} and ImageReward \cite{ImageReward}.
The fidelity reflects both low-level statistical naturalness and high-level perceptual quality aligned with human preferences.
Similar to \cite{DHY-BRISQUE}, we evaluate it as the normalized average of the scores obtained by the Blind/Referenceless Image Spatial Quality Evaluator (BRISQUE) \cite{BRISQUE} and the Multi-Scale Image Quality Transformer (MUSIQ) \cite{MUSIQ}.

Fig. \ref{fig:3-Generated-Examples} illustrates the experimental results under different guidance scales, averaged over both prompt groups, which reveal that a guidance scale of $w_1 = 2.0$ achieves the best trade-off, where both the alignment and fidelity scores reach their highest values.
In contrast, a higher guidance scale $w_1 = 7.0$ leads to semantic overfitting, generating overly deterministic but less relevant content, whereas a lower guidance scale $w_1 = 1.0$ introduces excessive randomness, which diminishes semantic coherence.
Based on these empirical observations, we select $w_1=2.0$ as the optimal guidance scale for Stable Diffusion 3 Medium \cite{SD3}.
This setting achieves the joint optimum of alignment and fidelity, thereby significantly enhancing the overall performance of the HIS.

\subsection{Hybrid Inference Quality Metric (HIQM)}\label{sec:HIQM}
It is easy to see that in the proposed HIS, the quality of the generated content is closely related to two key factors: (1) the semantic similarity between the public and personal prompts, and (2) the number of common inference steps.
Both factors significantly influence the semantic consistency of the outputs.

To systematically analyze these influences, we conduct a series of empirical experiments on an Ubuntu 22.04 system equipped with an Intel Core Ultra 9 285K CPU and two NVIDIA RTX 4090D GPUs.
Based on the experimental results, we formulate a novel metric, the HIQM, to quantify the quality of generated images under different conditions.
Fig. \ref{fig:5-Fitting-Function} presents the experimental procedure, which consists of four steps:

\subsubsection{Prompt Similarity Calculation (Step 1)}
To examine the generalization of the metric across prompt distributions, we construct two independent groups of representative prompts. The first group is synthesized by advanced LLMs \cite{Llama}, while the second group is sampled from real user prompts in the DiffusionDB dataset \cite{DiffusionDB}. Within each group, the prompts are randomly organized into public-personal prompt pairs.
The semantic similarity of each prompt pair is computed using a lightweight Sentence-Transformer model \cite{Sentence-Transformer}:
\begin{equation}\label{eq:Prompt-Similarity}\textstyle
	\phi = \text{Sim} \left( \text{emb}_{\text{S-T}} \left({\bf{p}}^{\text{pub}}\right), \text{emb}_{\text{S-T}} \left( {\bf{p}}^{\text{per}} \right) \right),
\end{equation}
where $\text{Sim}(\cdot, \cdot)$ and $\text{emb}_{\text{S-T}}(\cdot)$ are the cosine similarity and prompt embedding functions, respectively.
Here, the similarity is measured directly in the text-embedding space, as the Sentence-Transformer is dedicated to sentence-level semantic matching and is far more lightweight than image-text models such as CLIP.

\subsubsection{Image Generation (Step 2)}
To further examine the generalization across models, we adopt two representative GDMs, namely Stable Diffusion 3 Medium \cite{SD3} and FLUX.1-dev \cite{FLUX}, to generate images via the HIS with different  common inference steps.
In total, we generate $282$K images across the two models and the two prompt groups, each with a resolution of $1024 \times 1024$, allowing for robust statistical analysis.
\zxxy{We assess image quality in terms of two aspects: alignment and fidelity.}
From the generated images, we find that the common inference steps primarily affect the alignment of the generated images (with the personal prompt), and  the extent of this impact is closely related to the similarity of the public and personal prompts.
More specifically, (i) as the number of common inference steps increases, the alignment of the generated images gradually decreases, while the fidelity of the generated images remains consistently satisfactory;
(ii) with a higher prompt similarity, the alignment decreases at a slower rate as the number of common inference steps increases.
\zxxy{Based on these observations, we adopt alignment as the image-quality measure for the subsequent analysis and optimization.}

\begin{figure}[t]
	\centering
	\includegraphics[width=3.35in]{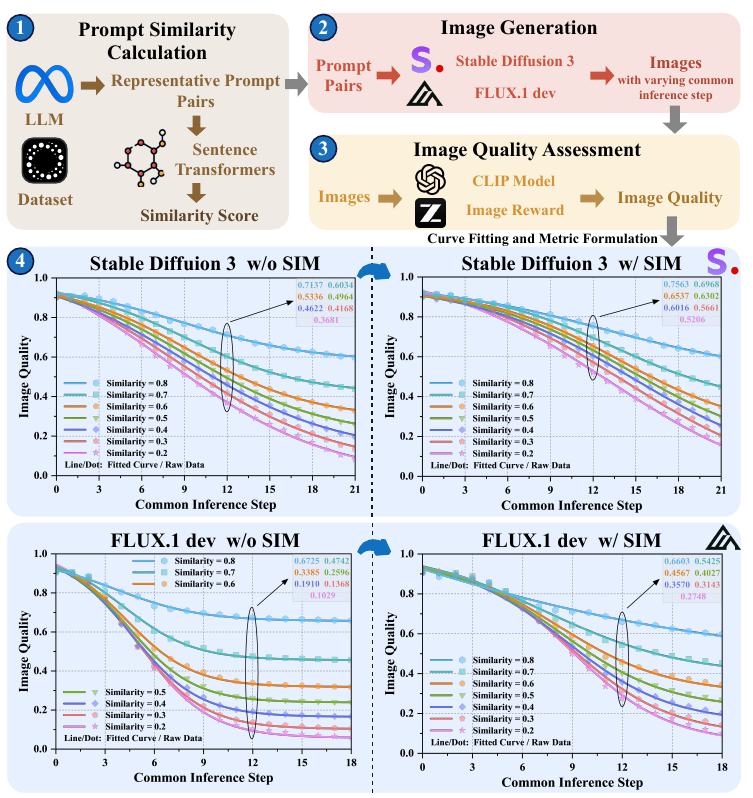}
	\vspace{-3mm}
	\caption{An overview of the experimental process and the corresponding fitted curves for Stable Diffusion 3 Medium and FLUX.1-dev, each with and without the SIM, based on real data averaged over the two prompt groups. \zxxy{The prompt similarity scores are rounded to the nearest 0.1 for statistical aggregation.}}
	\label{fig:5-Fitting-Function}
	\vspace{1mm}
\end{figure}

\subsubsection{Image Quality Assessment (Step 3)}
\zxxy{We quantify the image quality as the average of the normalized scores produced by the pre-trained ViT-L/14@336px CLIP model \cite{CLIP} and ImageReward \cite{ImageReward}:}
\begin{equation}\textstyle \label{eq:image-quality}
	{S_{\text{Quality}} = \tfrac{1}{2} \left( \tilde{S}_{\text{CLIP}} + \tilde{S}_{\text{IR}} \right),}
\end{equation}
where $\tilde{S}_{\text{CLIP}}$ and $\tilde{S}_{\text{IR}}$ denote the normalized scores of CLIP and ImageReward, respectively. In particular, the CLIP score is computed as $S_{\text{CLIP}} = \text{Sim} ( \text{emb}_{\text{CLIP}} ( {\bf{p}}^{\text{per}} ), \text{emb}_{\text{CLIP}} ( {\bf{I}} ) )$, where $\text{emb}_{\text{CLIP}}(\cdot)$ is the CLIP embedding function and ${\bf{I}}$ is the generated image at a given number of common inference steps. The ImageReward score is given by $S_{\text{IR}} = \text{IR} ( {\bf{p}}^{\text{per}}, {\bf{I}} )$, where $\text{IR}(\cdot, \cdot)$ denotes a reward model trained on large-scale human preference annotations. Unlike the CLIP score, which measures embedding-level cosine similarity, the ImageReward score reflects human-perceived prompt-image consistency, thereby providing a complementary assessment of alignment.

\subsubsection{Curve Fitting and Metric Formulation (Step 4)}
Fig. \ref{fig:5-Fitting-Function} (Part 4) illustrates the image quality under different common inference steps and prompt similarities.
We can see that as the  common inference steps increase, image quality initially declines gently, then deteriorates  rapidly, and eventually stabilizes.
To characterize the   relationship between the image quality and common inference steps, we use a modified sigmoid function to fit the quality-step curve.
That is, the HIQM function can be defined as follows:
\begin{equation}\textstyle \label{eq:HIQM}
	\hat{Q}(x) = \frac{L^{\phi,[\iota]}}{1+\exp\left(K^{\phi,[\iota]} \cdot \left(x - B^{\phi,[\iota]} \right) \right)} + G^{\phi,[\iota]},
\end{equation}
where $x$ is the number of common inference steps, and $L^{\phi,[\iota]}$, $K^{\phi,[\iota]}$, $B^{\phi,[\iota]}$, $G^{\phi,[\iota]}$ are hyper-coefficients related to prompt similarity $\phi$ and module indicator $\iota$ ($\iota=1$ if the SIM is enabled, $\iota=0$ otherwise).

Specifically, from Fig. \ref{fig:5-Fitting-Function} (Part 4), we can  find that as $x$ increases, the image quality gradually deteriorates, with the decline becoming more pronounced at lower prompt similarity.
That is, as the prompt similarity increases, the fitted curve becomes flatter and less abrupt, implying that images at higher similarity  maintain better quality.
Furthermore, when the SIM is enabled, the entire curve shifts upward and flattens further, implying that the SIM can effectively enhance the inference process by maintaining or improving image quality throughout the inference steps.
Notably, the same qualitative behavior is consistently observed for both Stable Diffusion 3 Medium and FLUX.1-dev, as well as for both the LLM-generated and DiffusionDB prompt groups. This indicates that the proposed HIQM formulation generalizes well across different GDMs and prompt distributions, rather than being tailored to a specific model or dataset.
Across all settings and similarity levels, the modified sigmoid attains an excellent fit, with $R^2 \geq 0.998$ and RMSE $\leq 0.013$ in the worst case.
In practice, these coefficients are fitted offline only once, and the quality of any UE is then obtained by evaluating the HIQM at its prompt similarity $\phi$, without any runtime fitting.

In the following, we will present the MEC-enabled AIGC network scenario, and formulate the joint user clustering, model inferencing, and resource allocation problem.


\section{System Model}\label{sec:SystemModel}

\subsection{Network Model}
\begin{figure}[t]
	\centering
	\includegraphics[width=3.35in]{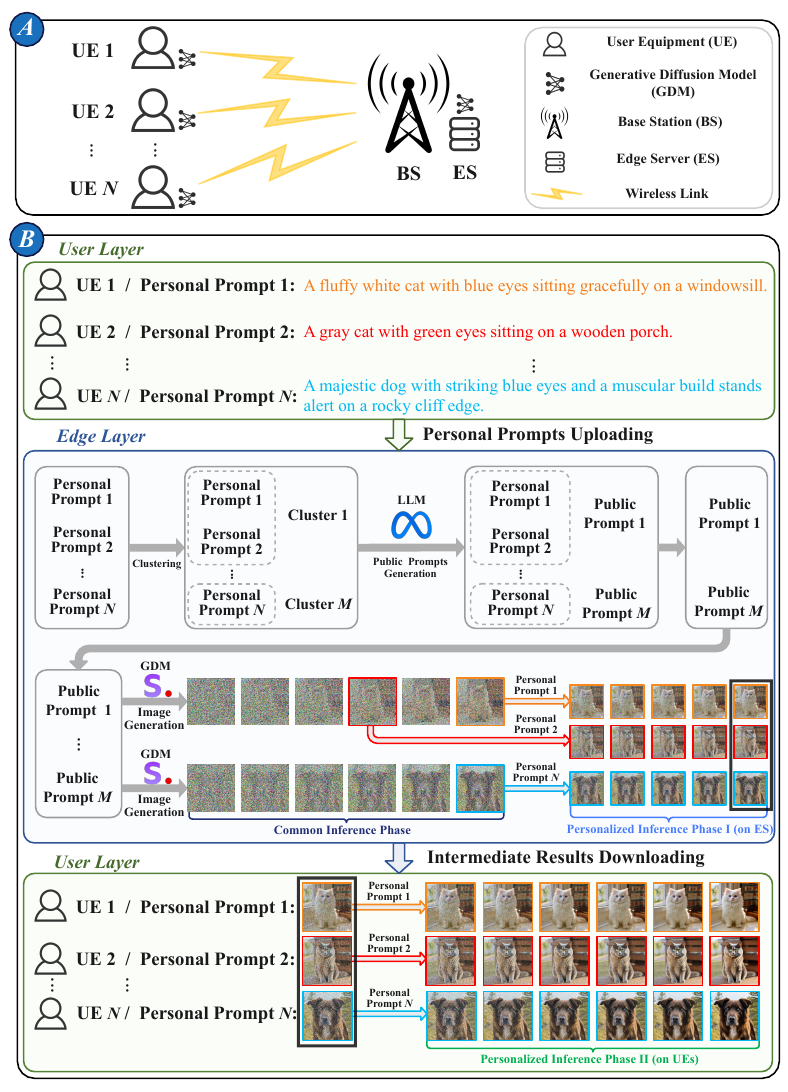}
	\vspace{-4mm}
	\caption{\textbf{Part A:} Illustration of an MEC network. \textbf{Part B:} Workflow of the HIS for MEC-enabled AIGC services.}
	\label{fig:6-System-Model}
	\vspace{1.5mm}
\end{figure}

As shown in Fig. \ref{fig:6-System-Model} (Part A), we consider an MEC network, which consists of a set $\mathcal{N} \triangleq \{ 1, \cdots, N \}$ of $N$ UEs (i.e., AIGC service requesters), each with limited computational resources, and one Base Station (BS) equipped with an ES that offers powerful computational resources to support AIGC services.
UEs request AIGC services to generate desired content.
With the advancement of model compression techniques, GDMs can be deployed on both the ES and UEs.

Fig. \ref{fig:6-System-Model} (Part B) illustrates the detailed workflow of the HIS for MEC-enabled AIGC services.
Specifically, at the user layer, UEs initiate their personal prompts, which are uploaded to the ES at the edge layer for further processing.
At the edge layer, the ES groups UEs with similar prompts into a \emph{cluster}, after which the LLM generates a corresponding public prompt for each cluster.
Subsequently, the HIS performs the following inference operations:
\begin{itemize}
	\item \textit{Common Inference Phase:} The GDM (on the ES) generates common intermediate results for each cluster based on the public prompt. These results will be shared by all UEs in the same cluster for later personalized inference.
	\item \textit{Personalized Inference Phase I (on ES):} The GDM (on the ES) selects an appropriate common intermediate result for each UE, and injects it (along with the UE's personal prompt) into the model to generate the personalized intermediate result for each UE. This result is then sent to the UE for subsequent inference.
	\item \textit{Personalized Inference Phase II (on UEs):} The GDM (on the UE) injects the received intermediate result into the model to generate the final result.
\end{itemize}

\subsection{Hybrid Inference Scheme}
\emph{1) User Clustering:}
As illustrated in Fig. \ref{fig:6-System-Model} (Part B), each UE $n \in \mathcal{N}$ is associated with a personal prompt ${\bf{p}}_n^{\text{per}}$.
Let ${\bf{P}}^{\text{per}} = \{ {\bf{p}}_n^{\text{per}} | n \in \mathcal{N}\}$ denote the set of all UEs' personal prompts.
The ES organizes all UEs into $M$ distinct clusters, denoted as $\mathcal{M} \triangleq \{ 1,\cdots, M \}$.
We introduce a binary clustering decision $\gamma_{n,m}\in \{0, 1\}$, with $\gamma_{n,m}=1$ indicating that UE $n$ is assigned to cluster $m$, and $\gamma_{n,m}=0$ otherwise.
For clarity, we use $\mathcal{N}_m$ to denote the set of UEs in cluster $m$.

After user clustering, we adopt a lightweight fine-tuned LLM (i.e., Qwen2.5-3B \cite{Qwen}) to extract the general semantic characteristics of prompts in each cluster, and generate a high-level, generalized public prompt\footnote{As a general task readily handled by a lightweight LLM, public prompt generation incurs negligible costs, and is therefore excluded from the problem formulation.} ${\bf{p}}_m^{\text{pub}}$ for each cluster $m$.

\begin{figure}[t]
	\centering
	\includegraphics[width=3.3in]{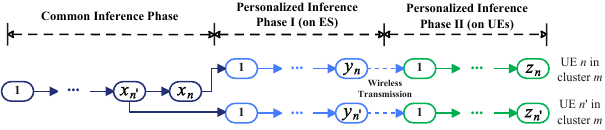}
	\vspace{-3mm}
	\caption{Illustration of the workflow of the HIS, where UE $n$ and $n'$ belong to the same cluster $m$.}
	\label{fig:7-HIS-Process}
	\vspace{1mm}
\end{figure}

\emph{2) Hybrid Inference:}
Fig. \ref{fig:7-HIS-Process} illustrates the detailed workflow of the HIS for each cluster $m$, where $x_n  $ denotes the common inference steps for UE $n$, $y_n  $ and $z_n $ denote the personalized inference steps on the ES and on UE $n$, respectively.
Specifically, in the common inference phase, the GDM uses the public prompt ${\bf{p}}_m^{\text{pub}}$ to perform $x_m^{\max} \triangleq \max\{ x_n, n \in \mathcal{N}_m \}$ inference steps, and generate a series of common intermediate results.
In personalized inference phase I (on ES), the GDM selects the corresponding intermediate result for each UE $n$, and performs $y_n$ inference steps to generate the personalized intermediate result for UE $n$.
In the personalized inference phase II (on UEs), the GDM on UE $n$ performs the remaining $z_n$ inference steps to generate the final result.

\subsection{Service Delay and Energy Consumption}
Now we calculate the service delay and energy consumption in each inference phase.

In the common inference phase, the GDM on the ES performs $x_m^{\max}$ inference steps for each cluster $m$.
Thus, the computation delay for cluster $m$ can be calculated as:
\begin{equation}\textstyle
	T_m^{\text{C}} = \frac{x_m^{\max} \xi^{\text{ES}}}{f_m^{\text{ES}}},
\end{equation}
where $\xi^{\text{ES}}$ is the number of Floating Point Operations (FLOPs) required for each inference step on the ES, and $f_m^{\text{ES}}$ is the number of FLOPs per Second (FLOPS) allocated to cluster $m$ on the GPU at the ES.

We adopt the GPU Streaming Multiprocessor (SM) partitioning strategy that enables precise control over parallel computational resources.
According to \cite{GPU-Sharing}, the spatial sharing technique enables concurrent tasks to share a single GPU efficiently while sustaining high utilization.
In particular, it partitions the GPU by assigning disjoint subsets of SMs to different tasks.
Based on this, $f_m^{\text{ES}}$ can be calculated as:
\begin{equation}\label{eq:GPU-FLOPs}\textstyle
	f_m^{\text{ES}} =  v_m^{\text{ES}} F^{\text{ES}} D^{\text{ES}} \rho^{\text{ES}},
\end{equation}
where $v_m^{\text{ES}}$ denotes the SMs assigned to cluster $m$, $F^{\text{ES}}$ is the computing frequency, $D^{\text{ES}}$ is the per-cycle per-SM FLOPs determined by the data and GPU type, and $\rho^{\text{ES}}$ is a discount factor reflecting non-computational bottlenecks.~~~~

Let $P_m^{\text{ES}}$ be the operating power of the ES GPU serving cluster $m$.
According to \cite{GPU-Power}, this power scales linearly with the number of active SMs and cubically with the computing frequency.
Thus, $P_m^{\text{ES}}$ can be calculated as:
\begin{equation}\textstyle
	P_m^{\text{ES}} = \kappa^{\text{ES}} \left( \mu_1^{\text{ES}} v_m^{\text{ES}} + \mu_2^{\text{ES}} \right) \left( F^{\text{ES}} \right)^3,
\end{equation}
where $\kappa^{\text{ES}}$ captures the chip-architecture-dependent coefficient, $\mu_1^{\text{ES}}$ quantifies the per-SM power growth, and $\mu_2^{\text{ES}}$ accounts for the baseline power.
Based on the above, the energy consumption in this phase can be calculated as:
\begin{equation}\label{eq:Energy-Consumption}\textstyle
	E_m^{\text{C}} = P_m^{\text{ES}} T_m^{\text{C}} = \frac{\kappa^{\text{ES}} x_m^{\max} \xi^{\text{ES}} (F^{\text{ES}})^2}{D^{\text{ES}} \rho^{\text{ES}}} \left( \mu_1^{\text{ES}} + \frac{\mu_2^{\text{ES}}}{v_m^{\text{ES}}}  \right).
\end{equation}

In the personalized inference phase I (on ES), the GDM on the ES performs $y_n$ inference steps for each UE $n$. Thus, the total computation delay  for all UEs in cluster $m$ is:\footnote{Here, we define the computation delay in this phase based on the worst-case estimation, i.e., the total time required for completing all personalized inference tasks in the cluster. As these tasks share the same GPU partition, this total time upper-bounds the ES-side waiting time of each UE.}
\begin{equation}\textstyle
	T_m^{\text{P-E}} = \sum_{n \in \mathcal{N}_m} \frac{y_n \xi^{\text{ES}}}{f_m^{\text{ES}}} = \sum_{n \in \mathcal{N}_m} \frac{y_n \xi^{\text{ES}}}{v_m^{\text{ES}} F^{\text{ES}} D^{\text{ES}} \rho^{\text{ES}}}.
\end{equation}
Similar to Eq. (\ref{eq:Energy-Consumption}), the energy consumption   can be calculated as:
\begin{equation}\textstyle
	E_m^{\text{P-E}} = \sum_{n \in \mathcal{N}_m}\frac{\kappa^{\text{ES}} y_n \xi^{\text{ES}} (F^{\text{ES}})^2}{D^{\text{ES}} \rho^{\text{ES}}} \left( \mu_1^{\text{ES}} + \frac{\mu_2^{\text{ES}}}{v_m^{\text{ES}}}  \right).
\end{equation}

In the personalized inference phase II (on UEs), UE $n$ downloads the personalized intermediate result from the ES via wireless links, after which the GDM on UE $n$ performs the remaining $z_n$ inference steps to generate the final result.
The wireless transmission delay and energy consumption can be calculated as:
\begin{equation}\textstyle
	T_n^{\text{T}} = \frac{n_{\text{h}} n_{\text{w}} n_{\text{c}} \varrho}{r_n \tau}, \quad E_n^{\text{T}} = \frac{p_n n_{\text{h}} n_{\text{w}} n_{\text{c}} \varrho}{r_n \tau},
\end{equation}
where $(n_{\text{h}} n_{\text{w}})$ is the total number of pixels in the latent space \cite{Latent-Space}, $n_{\text{c}}$ is the number of channels for each pixel, $\varrho$ is the number of bits required to represent the information per pixel per channel, $\tau$ is the compression ratio, and $r_n$ is the downlink transmission rate, which can be defined as:
\begin{equation}\textstyle
	r_n = b_n \log_2 \left( 1 + \frac{p_ng_n}{\sigma_0^2 b_n} \right),
\end{equation}
where $b_n$ and $p_n$ are the bandwidth and downlink transmission power, respectively, $g_n$ is the channel gain, and $\sigma_0^2$ is the
noise power spectral density.
The computation delay on UE $n$ can be calculated as:
\begin{equation}\textstyle
	T_n^{\text{P-U}} = \frac{z_n \xi^{\text{UE}}}{f_n^{\text{UE}}},
\end{equation}
where $\xi^{\text{UE}}$ is the number of FLOPs required for each inference step on the UE, and $f_n^{\text{UE}}$ is the available FLOPS on UE $n$.
Similar to Eq. (\ref{eq:GPU-FLOPs}), $f_n^{\text{UE}}$ can be calculated as:
\begin{equation}\textstyle
	f_n^{\text{UE}} = v_n^{\text{UE}} F_n^{\text{UE}} D_n^{\text{UE}} \rho^{\text{UE}},
\end{equation}
where $v_n^{\text{UE}}$ is the available SMs on UE $n$, $F_n^{\text{UE}}$ is the computing frequency, $D_n^{\text{UE}}$ is the number of FLOPs per cycle per SM, and $\rho^{\text{UE}}$ is the discount factor.
Similar to Eq. (\ref{eq:Energy-Consumption}), the energy consumption in this phase can be calculated as:
\begin{equation}\textstyle
	E_n^{\text{P-U}} = \frac{\kappa^{\text{UE}} z_n \xi^{\text{UE}} (F_n^{\text{UE}})^2}{D_n^{\text{UE}} \rho^{\text{UE}}} \left( \mu_1^{\text{UE}} + \frac{\mu_2^{\text{UE}}}{v_n^{\text{UE}}}  \right),
\end{equation}
where $\kappa^{\text{UE}}$, $\mu_1^{\text{UE}}$, and $\mu_2^{\text{UE}}$ are defined analogously.

Based on the above, the service delay of each UE $n$, denoted by $T_n$, can be calculated as:
\begin{equation}\textstyle
	T_n =  \sum_{m=1}^M \gamma_{n,m} \left( T_m^{\text{C}} + T_m^{\text{P-E}}\right) + T_n^{\text{T}} + T_n^{\text{P-U}}.
\end{equation}
Here, $T_n$ denotes the worst-case end-to-end service delay of UE $n$. The terms $T_m^{\text{C}}$ and $T_m^{\text{P-E}}$ are the common-phase delay and the worst-case ES-side waiting time shared by all UEs in cluster $m$, whereas $T_n^{\text{T}}$ and $T_n^{\text{P-U}}$ are the transmission and local computation delays specific to UE $n$.
The energy consumption of the ES for each cluster $m$, denoted by $E_m$, and the energy consumption of each UE $n$, denoted by $E_n$, can be calculated as:
\begin{equation}\textstyle
	E_m = E_m^{\text{C}} +E_m^{\text{P-E}}, \quad E_n = E_n^{\text{T}} + E_n^{\text{P-U}}.
\end{equation}

As shown in Section \ref{sec:HIS}, the service quality of each UE $n$, denoted by $Q_n$, can be evaluated by the HIQM function   defined in Eq. \eqref{eq:HIQM}.
Then, the total service quality $Q$, service delay $T$, and energy consumption $E$ can be calculated as:
\begin{equation}\label{eq:All-Service-Quality-Delay}\textstyle
	Q = \sum_{n=1}^{N} Q_n, \quad T = \sum_{n=1}^{N} T_n,
\end{equation}
\begin{equation}\label{eq:All-Energy-Consumption}\textstyle
	E = \sum_{m=1}^M E_m + \sum_{n=1}^N E_n.
\end{equation}
To balance the trade-off among service quality, service delay, and energy consumption, we define the \emph{network utility} as a weighted sum of these factors:
\begin{equation}\textstyle
	U \triangleq \varpi_1 Q - \varpi_2 T - \varpi_3 E,
	\vspace{-1mm}
\end{equation}
where $\varpi_1$, $\varpi_2$, and $\varpi_3$ are weight factors.

\vspace{-1mm}

\subsection{Problem Discussion}
Based on the above, we can find that the service quality, service delay, and energy consumption are all closely related to user clustering decisions (i.e., $\gamma_{n,m}$), model inferencing decisions (i.e., $x_n, y_n, z_n$), and resource allocation decisions (i.e., $v_m^{\text{ES}}$).
Therefore, in this work, we focus on the joint optimization of user clustering, model inferencing, and resource allocation to optimize the network utility   in the HIS.
These three types of decisions are tightly coupled and inherently sequential: user clustering determines the cluster composition and the resulting prompt similarities, upon which model inferencing and resource allocation depend. Following this structure, we decompose the joint problem into two subproblems.
In what follows, we formulate and solve the user clustering problem in Section \ref{sec:OptimizationI}, and subsequently the joint model inferencing and resource allocation problem in \mbox{Section \ref{sec:OptimizationII}}.
\section{Optimization of User Clustering}\label{sec:OptimizationI}
In this section, we study the user clustering problem.
As shown in Fig. \ref{fig:5-Fitting-Function}, the quality of the generated image increases with the prompt similarity.
Motivated by this, our goal is  to group UEs with similar prompts into the same cluster, such that the personal prompt of each UE is well aligned with the public prompt of the corresponding cluster.

Formally, we denote the clustering decisions with a matrix ${\bf{\Gamma}} \triangleq [\gamma_{n,m}]_{n \in \mathcal{N}, m \in \mathcal{M}}$, where $\gamma_{n,m}\in\{0,1\}$ denotes whether UE $n$ is assigned to cluster $m$.
Let ${\bf{p}}_n^{\text{per}}$ denote the personal prompt of UE $n$, and ${\bf{p}}_m^{\text{pub}}$ denote the public prompt extracted from cluster $m$.
Let $\phi_{n,m}$ denote the public-personal prompt similarity between the public prompt ${\bf{p}}_m^{\text{pub}}$ of cluster $m$ and the personal prompt ${\bf{p}}_n^{\text{per}}$ of UE $n$, which is derived from Eq. \eqref{eq:Prompt-Similarity}.
It is notable that  both the public prompt ${\bf{p}}_m^{\text{pub}}$ and the public-personal prompt similarity $\phi_{n,m}$ are dependent on the clustering decisions ${\bf{\Gamma}}$, and thus can be expressed as ${\bf{p}}_m^{\text{pub}} ({\bf{\Gamma}})$ and $\phi_{n,m}({\bf{\Gamma}})$, respectively.

Based on the above, we can formulate the user clustering optimization problem as follows:
\begin{subequations}
	\vspace{0mm}
	\begin{align}
		\mathbb{P}_{\text{I}}^{(1)}:~~
		\max\limits_{
			\tiny
			\begin{array}{c}
				\small
				{\bf{\Gamma}}
		\end{array}}~~
		&\textstyle
		\sum_{n=1}^N \sum_{m=1}^M  \gamma_{n,m} \phi_{n,m}({\bf{\Gamma}}) \\
		\text{s.t.} ~~~
		&\textstyle
		\gamma_{n,m} \in \{ 0, 1 \}, \quad \forall n \in \mathcal{N}, \ \forall m \in \mathcal{M}
		\label{eq:Constraint1-1}\\
		&\textstyle
		\sum_{m=1}^M \gamma_{n,m} = 1, \quad \forall n \in \mathcal{N}.
		\label{eq:Constraint1-2}
	\end{align}
\end{subequations}

\subsection{Problem Transformation}
\begin{figure}[t]
	\centering
	\includegraphics[width=3.2in]{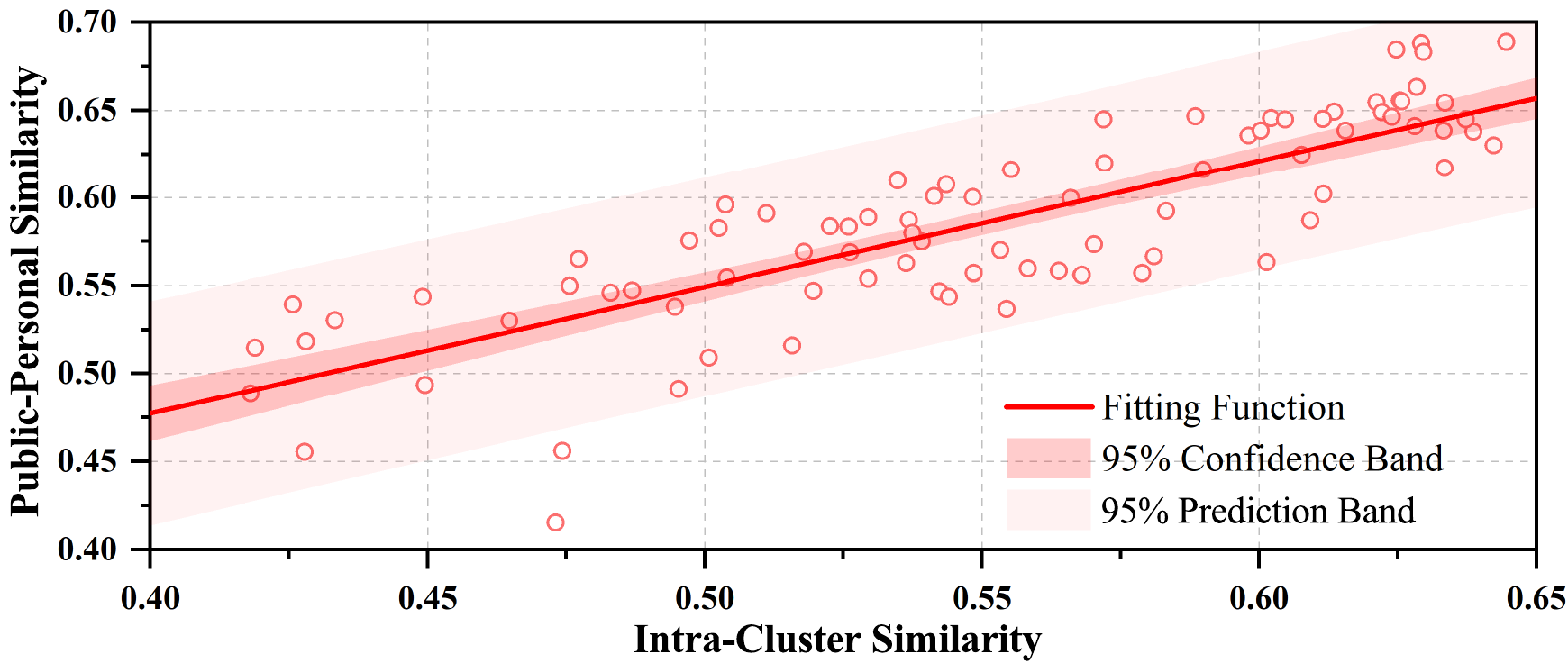}
	\vspace{-3mm}
	\caption{Relationship between two similarity metrics.}
	\label{fig:8-Personal-Public-Similarity}
	\vspace{-1mm}
\end{figure}

It is important to note that directly solving $\mathbb{P}_{\text{I}}^{(1)}$ is infeasible, due to the strong coupling between the public-personal prompt similarity $\phi_{n,m}$ and the clustering decisions ${\bf{\Gamma}}$, as well as the intractability in strictly characterizing this coupling.
To deal with this issue, we conduct empirical experiments to analyze the relationship between the public-personal similarity and the intra-cluster similarity.
\begin{enumerate}
\item \emph{Intra-Cluster Similarity:} The average similarity between personal prompts within the same cluster.
\item \emph{Public-Personal Similarity:} The average similarity between each personal prompt and its corresponding public prompt (i.e., the average value of $\phi_{n,m}$ in each cluster).
\end{enumerate}
Fig. \ref{fig:8-Personal-Public-Similarity} presents the experimental results, which reveal a high linear correlation between these two similarity metrics.
That is, as the intra-cluster similarity increases, the public-personal similarity also increases linearly, and vice versa.

Let $\hat{\phi}_{n,n'}$ denote the similarity between personal prompts of UEs $n$ and $n'$, i.e., ${\bf{p}}_n^{\text{per}}$ and ${\bf{p}}_{n'}^{\text{per}}$.
Clearly, $\hat{\phi}_{n,n'}$ is independent of the clustering decisions ${\bf{\Gamma}}$.
Based on the above observation, we can transform $\mathbb{P}_{\text{I}}^{(1)}$ into the following new problem $\mathbb{P}_{\text{I}}^{(2)}$, which aims to maximize the intra-cluster similarity.
\begin{subequations}
	\begin{align}
		\mathbb{P}_{\text{I}}^{(2)}:
		\max\limits_{
			\tiny
			\begin{array}{c}
				\small
				{\bf{\Gamma}}
		\end{array}}~~
		& \textstyle
		\sum_{m\in \mathcal{M}} \sum_{n \in \mathcal{N}} \sum_{n' \in \mathcal{N}} \gamma_{n,m} \gamma_{n',m} \hat{\phi}_{n,n'} \\
		\text{s.t.} ~~~
		&\textstyle
		\eqref{eq:Constraint1-1}-\eqref{eq:Constraint1-2}, \notag
	\end{align}
\end{subequations}
where $\gamma_{n,m} \gamma_{n',m} =1$ indicates that UEs $n$ and $n'$ are in the same cluster $m$.
Note that  $\mathbb{P}_{\text{I}}^{(2)}$ is \textit{non-convex} and challenging due to the product of    variables $\gamma_{n,m}$ and $\gamma_{n',m}$.
To this end, we introduce  auxiliary variables $\mathbf{\Lambda} \triangleq  [\lambda_{n,n',m}]_{n, n' \in \mathcal{N}, m \in \mathcal{M}}$ and transform $\mathbb{P}_{\text{I}}^{(2)}$ into an equivalent problem $\mathbb{P}_{\text{I}}^{(3)}$:
\begin{subequations}
	\vspace{-1mm}
	\begin{align}
		\mathbb{P}_{\text{I}}^{(3)}:
		\max\limits_{
			\tiny
			\begin{array}{c}
				{\bf{\Gamma}}, {\bf{\Lambda}}
		\end{array}}
		&\textstyle
		\sum_{m\in \mathcal{M}}  \sum_{n \in \mathcal{N}} \sum_{n' \in \mathcal{N}} \lambda_{n, n', m} \hat{\phi}_{n,n'} \\
		\text{s.t.} ~
		&\textstyle
		\eqref{eq:Constraint1-1}-\eqref{eq:Constraint1-2} \notag \\
		&\textstyle
		\lambda_{n,n',m} \in \{0,1\},\quad \forall n, \forall n', \forall m \\
		&\textstyle
		\lambda_{n,n',m} \leq \gamma_{n,m}, \quad \forall n, \forall n', \forall m \\
		&\textstyle
		\lambda_{n,n',m} \leq \gamma_{n',m}, \quad \forall n, \forall n', \forall m \\
		&\textstyle
		\lambda_{n,n',m} \geq \gamma_{n,m} + \gamma_{n',m} -1,\forall n, \forall n', \forall m
	\end{align}
\end{subequations}
Clearly, $\mathbb{P}_{\text{I}}^{(3)}$ is an integer linear programming problem, widely known to be NP-Hard.
While exact solutions can be obtained using algorithms such as branch-and-bound and cutting-plane methods, their computational complexity $\mathcal{O}(2^{N^2M})$ grows exponentially with the problem size, making them unsuitable for large-scale MEC networks.

\begin{algorithm}[t]\label{algorithm1}
	\caption{Semantic-Based Clustering Algorithm}
	Initialize each personal prompt as a cluster;\\
	Calculate the similarity matrix $\hat{\mathbf{\Phi}}$;\\
	\For{\textnormal{each iteration $k = 1, \cdots, N-M$}}{
		Calculate all cluster-wise similarities based on  \eqref{eq:Clustering-HAC};\\
		Merge two clusters with the highest similarity;\\
	}
\end{algorithm}

\subsection{Algorithm Design}
To effectively solve $\mathbb{P}_{\text{I}}^{(3)}$, we design an efficient semantic-based clustering algorithm based on the HAC \cite{HAC-Clustering}, as shown in Algorithm \ref{algorithm1}.
Specifically, we first compute the similarity matrix $\hat{\mathbf{\Phi}} \triangleq  [\hat{\phi}_{n,n'}]_{n,n' \in \mathcal{N}}$ using the lightweight Sentence-Transformer model.
Then, the algorithm begins by treating each UE (personal prompt) as an individual cluster.
In each iteration, the algorithm merges the two clusters with the highest cluster-wise similarity according to:
\begin{equation}\label{eq:Clustering-HAC}\textstyle
	\zxxy{\ell(m,m') = \frac{1}{|\mathcal{N}_m| |\mathcal{N}_{m'}|}\sum_{n \in \mathcal{N}_m}\sum_{n' \in \mathcal{N}_{m'}} \hat{\phi}_{n,n'}.}
\end{equation}
This process continues until a predefined cluster number $M$ is reached.
Intuitively, this algorithm enhances intra-cluster similarity by iteratively merging the most similar clusters.
With a complexity of $\mathcal{O}(N^2 \log N)$, it provides a powerful solution to $\mathbb{P}_{\text{I}}^{(3)}$ that can be executed efficiently on the ES.

\section{Optimization of Model Inferencing and Resource Allocation}\label{sec:OptimizationII}
Based on the user clustering results obtained from Algorithm \ref{algorithm1}, we study the joint model inferencing and resource allocation problem in this section.
For notational convenience, we introduce the following notations: ${\bf{X}} \triangleq [x_n]_{n \in \mathcal{N}}$, ${\bf{Y}} \triangleq [y_n]_{n \in \mathcal{N}}$, ${\bf{Z}} \triangleq [z_n]_{n \in \mathcal{N}}$, and ${\bf{V}} \triangleq [v_m^{\text{ES}}]_{m \in \mathcal{M}}$.
Then, the joint model inferencing and resource allocation optimization problem can be formulated as:
\begin{subequations}
	\vspace{0mm}
	\begin{align}
		\mathbb{P}_{\text{II}}:
		\max\limits_{
			\tiny
			\begin{array}{c}
				\small
				{\bf{X}},{\bf{Y}},{\bf{Z}}, {\bf{V}}
		\end{array}}
		&\textstyle
	    U \triangleq	\varpi_1 Q - \varpi_2 T - \varpi_3 E
		\label{eq:Overall Optimization Problem}\\
		\text{s.t.} ~~~~
		&\textstyle
		x_{n} \in \{ 0, \cdots, x_{\max} \}, \quad \forall n \in \mathcal{N}
		\label{eq:Constraint1-3}\\
		&\textstyle
		y_{n} \in \{ y_{\min}, \cdots, y_{\max} \}, \quad \forall n \in \mathcal{N}
		\label{eq:Constraint2-3}\\
		&\textstyle
		z_{n} \in \{ z_{\min}, \cdots, z_{\max} \}, \quad \forall n \in \mathcal{N}
		\label{eq:Constraint3-3}\\
		&\textstyle
		\sum_{m=1}^M v_m^{\text{ES}} \leq v_{\max}^{\text{ES}}, v_m^{\text{ES}} \in \mathbb{Z}_+
		\label{eq:Constraint4-3}\\
		&\textstyle
		\zxxy{x_n + y_n + z_n = N_{\mathrm{step}}, \quad \forall n \in \mathcal{N}.}
		\label{eq:Constraint5-3}
	\end{align}
\end{subequations}
Note that constraint \eqref{eq:Constraint5-3} sets the total number of inference steps as a pre-determined constant. 
\zxxy{The value of $N_{\mathrm{step}}$ represents the total number of inference steps required to achieve satisfactory results.}

\subsection{Problem Decomposition}
Directly optimizing $\mathbb{P}_{\text{II}}$ is challenging due to the heterogeneous decision variables and the asynchronous decision-making process.
To streamline the optimization process, we decompose $\mathbb{P}_{\text{II}}$ into a \emph{user-wise} model inferencing problem $\mathbb{P}_{\text{II}}^{\text{A}}$ and a \emph{cluster-wise} resource allocation problem $\mathbb{P}_{\text{II}}^{\text{B}}$:
\begin{subequations}
	\begin{align}
		\mathbb{P}_{\text{II}}^{\text{A}}: 
		\max\limits_{
			\tiny
			\begin{array}{c}
				\small
				{\bf{X}},{\bf{Y}},{\bf{Z}}
		\end{array}}~
		&\textstyle
		U~~~~~
		\text{s.t.} ~\textstyle
		\eqref{eq:Constraint1-3}, \eqref{eq:Constraint2-3}, \eqref{eq:Constraint3-3}, \eqref{eq:Constraint5-3}. ~~\notag
	\end{align}
\end{subequations}
\begin{subequations}
	\vspace{-3mm}
	\begin{align}
		~\mathbb{P}_{\text{II}}^{\text{B}}:~~
		\max\limits_{
			\tiny
			\begin{array}{c}
				\small
				{\bf{V}}
		\end{array}}~~~
		&\textstyle
		U~~~~~
		\text{s.t.} ~\textstyle
		\eqref{eq:Constraint4-3}.~~~~~~~~~~~~~~~~~~~~~~~~~ \notag
	\end{align}
	\vspace{-4mm}
\end{subequations}

To solve the above two sub-problems, we propose a two-tier Model Inferencing and Resource Allocation (MIRA) algorithm.
In the first tier, we design a \emph{learning-based} approach to make real-time model inferencing decisions for UEs.
In the second tier, we employ \emph{optimization-based} techniques to make optimal resource allocation decisions for clusters.
Next, we will provide the detailed algorithm designs.

\subsection{Learning-Based Model Inferencing}
In the first tier, we utilize the DRL algorithm to solve sub-problem $\mathbb{P}_{\text{II}}^{\text{A}}$.
Note that the state-action space grows exponentially with the number of agents (i.e., UEs).
This curse of dimensionality poses challenges to the centralized execution scheme.
On the other hand, since parts of the inference process are performed on the ES, the decision of each agent needs to be uploaded to the ES.
This interaction process introduces unnecessary communication overhead, posing challenges to the traditional decentralized execution scheme.~~~~~~~~~

To address the above challenges, we employ the IPPO approach\footnote{We adopt IPPO rather than centralized-critic methods such as MAPPO \cite{MAPPO} or MADDPG \cite{MADDPG}, since its per-agent critic depends only on local observations and thus scales with the number of UEs, while also suiting the discrete inference step decisions.} \cite{IPPO} and introduce the centralized control for the multi-agent system. 
Specifically, the ES acts as a centralized controller, hosting the actor and critic networks of all agents (UEs).
The actor network is responsible for selecting the action to be performed, while the critic network is used to evaluate the action by estimating the expected future rewards.

Next, we define the observation, action, and reward.

\emph{1) Observation:} The observation consists of two parts: resource status and prompt similarity.
Here, we define the observation of UE $n$ at decision point $i$ as:
\begin{equation}\textstyle
	{\bf{o}}_n^{(i)} = ( \phi_{n,m}^{(i)}, v_{\max}^{\text{ES}, (i)}, v_n^{\text{UE}, (i)}).
\end{equation}

\emph{2) Action:} The action consists of two parts: the number of common and personalized inference steps.
Here, we define the action of UE $n$ at decision point $i$ as:
\begin{equation}\textstyle
	{\bf{a}}_n^{(i)} = ( x_n^{(i)}, y_n^{(i)} ).
\end{equation}

\emph{3) Reward:}
To precisely capture the reward improvement resulting from different actions, the reward of UE $n$ at decision point $i$ is defined as:
\begin{equation}\textstyle
	\begin{aligned}
		&\textstyle
		r_n^{(i)}
		= \varpi_1 Q_n - \varpi_2 T_n - \varpi_3 E_n \\
		&\textstyle
		\quad \quad - \varpi_3 \sum_{m=1}^M \gamma_{n,m}
		\left(
		\tfrac{x_n E_m^{\text{C}}}{\sum_{n' \in \mathcal{N}_m} x_{n'}}
		+
		\tfrac{y_n E_m^{\text{P-E}}}{\sum_{n' \in \mathcal{N}_m} y_{n'}}
		\right).
	\end{aligned}
\end{equation}
Here, the ES energy $E_m$ of each cluster is split among its UEs in proportion to their inference steps.
Thus, we have $U^{(i)} = \sum_{n=1}^N r_n^{(i)} = \varpi_1 Q - \varpi_2 T - \varpi_3 E$.
To improve stability and accelerate convergence, we introduce reward normalization to adjust the scale of rewards.
Specifically, reward normalization dynamically maintains the mean and standard deviation of all observed rewards for each agent, normalizing the current reward by subtracting the mean and dividing by the standard deviation.
This ensures that rewards are centered around zero with unit variance, mitigating the negative effects of large or small rewards on the training process.

\begin{figure}[t]
	\centering
	\includegraphics[width=3.4in]{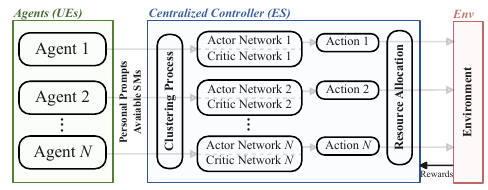}
	\vspace{-3mm}
	\caption{Decision process of the learning-based model inferencing algorithm.}
	\label{fig:9-MIRA}
	\vspace{1mm}
\end{figure}

As shown in Fig. \ref{fig:9-MIRA}, at each decision point, the ES obtains individual observation ${\bf{o}}_n^{(i)}$ and chooses corresponding action ${\bf{a}}_n^{(i)}$ for all UEs in a parallel way.
Once all content has been generated, the ES obtains rewards and next individual observations, and stores the experiences in corresponding replay buffers.
\zxxy{Let $\rho_n^{(d,i)} \triangleq \pi_{{\bf{\Theta}}_n}({\bf{a}}_{n}^{(d,i)} \mid {\bf{o}}_{n}^{(d,i)}) / \pi_{{\bf{\Theta}}_n^{\text{old}}}({\bf{a}}_{n}^{(d,i)} \mid {\bf{o}}_{n}^{(d,i)})$ denote the probability ratio.}
\zxxy{The actor network can then be updated by maximizing the following objective:}
\begin{equation}\textstyle
	\begin{split}
		\small
		&\textstyle
		\mathcal{L}_n^{\text{A}_1}
		= \arg\max\limits_{{\bf{\Theta}}_n}
		\bigg\{
		\frac{1}{DI} \sum_{d=1}^{D} \sum_{i=1}^{I}
		\min \bigg[
		\rho_n^{(d,i)} \hat{A}_n^{(d,i)}, \\
		&\textstyle
		\qquad\qquad
		\mathrm{trun}\!\left(
		\rho_n^{(d,i)}, 1-\varepsilon, 1+\varepsilon
		\right)
		\hat{A}_n^{(d,i)}
		\bigg]
		\bigg\}.
	\end{split}
\end{equation}
where $\pi_{{\bf{\Theta}}_n}$ is the actor network for UE $n$, $D$ is the batch size, $\varepsilon$ is the coefficient of the truncation function $\mathrm{trun}(\cdot)$, and ${\hat{A}}_{n}^{(d, i)}$ is the generalized advantage estimation. 

To improve the algorithm performance, we further introduce batch advantage normalization \cite{Advantage-Normalization} to normalize the advantages by subtracting the mean and dividing by the standard deviation of the entire batch.
Furthermore, to balance exploration and exploitation, we introduce the policy entropy term for the loss function, which can be defined as:
\begin{equation}\textstyle
	\mathcal{L}_n^{\text{A}_2} =  \arg\max\limits_{{\bf{\Theta}}_n} \left\{
	\frac{1}{DI} \sum_{d=1}^D \sum_{i=1}^I \mathbb{E}_{\bf{a}} \left[-\log \pi_{{\bf{\Theta}}_n} ({\bf{a}} | {\bf{o}}_{n}^{(d, i)}) \right] \right\}.
\end{equation}
Thus, the loss function of the actor network is as follows:
\begin{equation}\label{eq:Actor-Loss} \textstyle
	\mathcal{L}_n^{\text{A}} = \mathcal{L}_n^{\text{A}_1} + \eta	\mathcal{L}_n^{\text{A}_2},
\end{equation}
where $\eta$ is a coefficient controlling the strength of the entropy regularization.
Finally, the critic network can be updated according to the following loss function:
\begin{equation}\label{eq:Critic-Loss}
	\vspace{-1mm}
	\begin{split}
		\mathcal{L}_n^{\text{C}} =  &
		\textstyle
		\arg\min\limits_{{\bf{\Phi}}_n} \left\{
		\frac{1}{DI} \sum_{d=1}^D \sum_{i=1}^I \max \left\{
		\left( V_{{\bf{\Phi}}_n}({\bf{o}}_n^{(d,i)})
		\right. \right. \right.\\
		&\textstyle
		\left. \left. \left.
		- {\hat{R}}_{n}^{(d,i)} \right)^2,
		\left( \mathrm{trun}\left(V_{{\bf{\Phi}}_n}({\bf{o}}_n^{(d,i)}),
		V_{{\bf{\Phi}}_n^{\text{old}}}({\bf{o}}_n^{(d,i)})-\epsilon,
		\right. \right. \right. \right.\\
		&\textstyle
		\ \left. \left. \left. \left.
		V_{{\bf{\Phi}}_n^{\text{old}}}({\bf{o}}_n^{(d,i)})+\epsilon \right)
		-{\hat{R}}_{n}^{(d,i)} \right)^2
		\right\}
		\right\},
	\end{split}
	\raisetag{18pt}
\end{equation}
where $V_{{\bf{\Phi}}_n}$ is the critic network for UE $n$, $\epsilon$ is the coefficient of the truncation function, and ${\hat{R}}_{n}^{(d,i)}$ is the reward-to-go.

\subsection{Optimization-Based Resource Allocation}
Once the model inferencing decisions (i.e., $\bf{X}, \bf{Y}, \bf{Z}$) are determined, the sub-problem $\mathbb{P}_{\text{II}}^{\text{B}}$ can be written as:
\begin{subequations}
	\begin{align}
		\mathbb{P}_{\text{II}}^{\text{B}}:~~
		\min\limits_{
			\tiny
			\begin{array}{c}
				\small
				{\bf{V}}
		\end{array}}~~
		\textstyle
		\sum_{m=1}^M \left( \frac{\alpha_m^{(1)} + \alpha_m^{(2)}}{v_m^{\text{ES}}} \right)~~~
		\text{s.t.}~~
		\textstyle
		\eqref{eq:Constraint4-3}, \notag
	\end{align}
\end{subequations}
where $\alpha_m^{(1)} = \sum_{n=1}^N \frac{\left( x_m^{\max} + \sum_{n' \in \mathcal{N}_m} y_{n'}  \right) \gamma_{n,m} \varpi_2 \xi^{\text{ES}}}{F^{\text{ES}} D^{\text{ES}} \rho^{\text{ES}}}$ and $\alpha_m^{(2)} = \frac{ \left( x_m^{\max} + \sum_{n \in \mathcal{N}_m} y_n \right) \varpi_3 \kappa^{\text{ES}} \xi^{\text{ES}} (F^{\text{ES}})^2 \mu_2^{\text{ES}} }{D^{\text{ES}} \rho^{\text{ES}}}$.

Next, we provide the closed-form solution to $\mathbb{P}_{\text{II}}^{\text{B}}$ by relaxing the decision variables.
\begin{theorem}
	The relaxed sub-problem $\mathbb{P}_{\text{II}}^{\text{B}}$ is convex, and its optimal solution $(v_m^{\text{ES}})^{\star}$ is given by\footnote{In practice, $(v_m^{\text{ES}})^{\star}$ can be rounded down with minimal performance loss.}:
	\begin{equation}\label{eq:Optimal-V}\textstyle
		(v_m^{\text{ES}})^{\star} = v_{\max}^{\text{ES}} \frac{\sqrt{\alpha_m^{(1)} + \alpha_m^{(2)}}}{\sum_{m \in \mathcal{M}} \sqrt{\alpha_m^{(1)} + \alpha_m^{(2)}}} ,\quad \forall m \in \mathcal{M}.
	\end{equation}
\end{theorem}

\begin{proof}
	The detailed proof is provided in Appendix.
\end{proof}

\subsection{Algorithm Description}
Based on the above, we summarize the MIRA algorithm in Algorithm \ref{algorithm2}.
At each decision point, each UE uploads its personal prompt to the ES, after which the ES groups UEs into clusters according to Algorithm \ref{algorithm1}.
Then, the ES collects individual observations and chooses corresponding actions for all UEs (lines 6-7).
Next, the ES decides computational resource allocation based on Eq. \eqref{eq:Optimal-V} and performs the common inference phase and personalized inference phase I based on all actions (lines 8-9).
After wireless transmission, the personalized inference phase II is performed on the UE based on Eq. \eqref{eq:Constraint5-3}.
Once all content has been generated, the ES stores the experiences in corresponding replay buffers (lines 10-11).
Finally, the actor and critic networks can be updated by Eq. \eqref{eq:Actor-Loss} and Eq. \eqref{eq:Critic-Loss} (lines 14-15).

\begin{algorithm}[h]\label{algorithm2}
	\caption{MIRA Algorithm}
	Initialize actor networks $\{\pi_{{\bf{\Theta}}_n}\}_{n \in \mathcal{N}}$, critic networks $\{V_{{\bf{\Phi}}_n}\}_{n \in \mathcal{N}}$, replay buffers $\{\mathcal{D}_n\}_{n \in \mathcal{N}}$, and learning parameters\;
	\For{\zxxy{\textnormal{each episode} $e = 1, \cdots, N_{\mathrm{ep}}$}}{
		\For{\textnormal{each decision point} $i = 1, \cdots, I$}{
			Each UE $n$ uploads its personal prompt and resource status to the ES\;
			The ES groups UEs into clusters according to Algorithm \ref{algorithm1}\;
			The ES collects observations $\{{\bf{o}}_n^{(i)}\}_{n \in \mathcal{N}}$ for UEs\;
			The ES selects actions $\{ {\bf{a}}_n^{(i)} \}_{n \in \mathcal{N}}$ for UEs based on actor networks $\{ \pi_{{\bf{\Theta}}_n}({\bf{a}}_{n}^{(i)} \mid {\bf{o}}_{n}^{(i)}) \}_{n \in \mathcal{N}}$\;
			The ES optimizes computational resource allocation $\{v_m^{\text{ES}}\}_{m \in \mathcal{M}}$ by  \eqref{eq:Optimal-V}\;
			Execute actions for all UEs simultaneously\;
			The ES obtains rewards $\{r_n^{(i)}\}_{n \in \mathcal{N}}$ and collects new observations $\{{\bf{o}}_n^{(i+1)}\}_{n \in \mathcal{N}}$ for UEs\;
			The ES stores experiences $\{({\bf{o}}_n^{(i)}, {\bf{a}}_n^{(i)}, r_n^{(i)}, {\bf{o}}_n^{(i+1)})\}_{n \in \mathcal{N}}$ in replay buffers $\{\mathcal{D}_n\}_{n \in \mathcal{N}}$\;
		}
	}
	\For{\textnormal{each training step}}{
		\For{\textnormal{each UE} $n = 1, \cdots, N$}{
			Update actor network $\pi_{{\bf{\Theta}}_n}$ by maximizing \eqref{eq:Actor-Loss}\;
			Update critic network $V_{{\bf{\Phi}}_n}$ by minimizing \eqref{eq:Critic-Loss}.
		}
	}
\end{algorithm}

\section{Performance Evaluation}\label{sec:PerformanceEvaluation}

\begin{figure*}[t]
	\centering
	\includegraphics[width=7in]{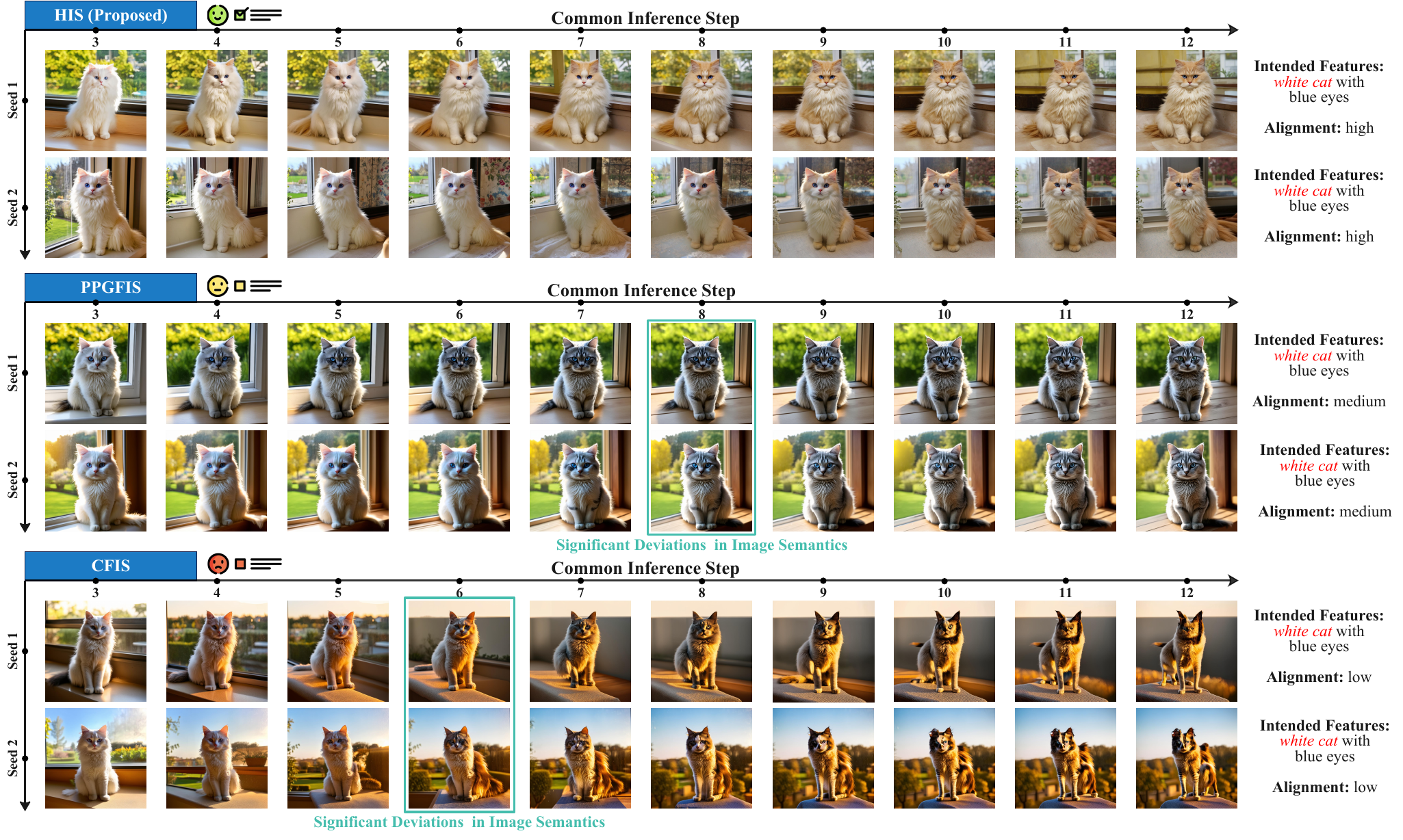}
	\vspace{-3mm}
	\caption{Images generated by the HIS (\textbf{Top}), PPGFIS (\textbf{Middle}), and CFIS (\textbf{Bottom}) under different common inference steps. All images are based on the same personal prompt: \textit{A fluffy white cat with blue eyes sitting gracefully on a windowsill.}}
	\label{fig:10-Case-Study} 
\end{figure*}

\subsection{Experimental Settings}
In the simulations, we consider a wireless network consisting of one ES and 50 UEs randomly distributed within a radius of 50 to 300 meters from the ES.
To simulate diverse generation requirements, the UEs are grouped into three distinct clusters.
To support advanced AIGC services, both the ES and UEs deploy the state-of-the-art Stable Diffusion 3 Medium model \cite{SD3}, which is configured with the Flow-Match-Euler-Discrete-Scheduler to ensure high-quality image generation.
\zxxy{The total number of inference steps is fixed at 28 steps (i.e., $N_{\mathrm{step}} = 28$) as suggested in \cite{SD3}.}
For the hardware setting, the ES is equipped with an industrial-grade NVIDIA A100 GPU cluster, with each GPU featuring 108 SMs for intensive processing.
Meanwhile, each UE is equipped with a consumer-grade RTX 3090 Ti GPU, each with 84 SMs for real-time processing.
Additional key environmental and algorithmic parameters are summarized in Table \ref{table:Parameters}.

\begin{table}[t]
	\vspace{-2.5mm}
	\caption{Default Parameter Settings}
	\label{table:Parameters}
	\vspace{-2mm}
	\centering
	\begin{tabular}{ll}
		\Xhline{1pt}
		Parameter & Value \\
		\Xhline{0.3pt}
		$N$, $M$, $x_{\max}$, $y_{\min}$, $y_{\max}$ & $50$, $3$, $16$, $0$, $11$ \\
		$\xi^{\text{ES}}$, $v_{\max}^{\text{ES}}$, $F^{\text{ES}}$ & $12$ TFLOPs, $[1026, 1674]$, $1410$ MHz \\
		$D^{\text{ES}}$, $\kappa^{\text{ES}}$, $\mu_1^{\text{ES}}$, $\mu_2^{\text{ES}}$ & $2048$, $10^{-27}$, $4$, $135$ \\
		\zxxy{$\xi^{\text{UE}}$, $v_n^{\text{UE}}$, $F_n^{\text{UE}}$} & \zxxy{$12$ TFLOPs, $[64, 84]$, $1860$ MHz} \\
		$D^{\text{UE}}$, $\kappa^{\text{UE}}$, $\mu_1^{\text{UE}}$, $\mu_2^{\text{UE}}$ & $1024$, $10^{-27}$, $4$, $135$ \\
		$n_{\text{h}}$, $n_{\text{w}}$, $n_{\text{c}}$, $\varrho$, $\tau$  & $128$, $128$, $16$, $16$, $2$ \\
		learning rate, hidden layers & $0.005$, $[256, 256, 256]$ \\
		optimizer, $D$, $\varepsilon$, $\epsilon$, $\eta$ & Adam, $128$, $0.2$, $0.2$, $0.002$ \\
		\Xhline{1pt}
	\end{tabular}
\end{table}

\subsection{The Effectiveness and Robustness of the HIS}
In this subsection, we present simulation results to demonstrate the effectiveness and robustness of our proposed HIS.
For comparisons, we introduce the following three inference schemes from the existing literature as benchmarks:
\begin{itemize}
	\vspace{0mm}
	\item \emph{\underline{P}ublic \underline{P}rompt \underline{G}eneration-\underline{F}ree \underline{I}nference \underline{S}cheme (\mbox{PPGFIS})} \cite{XGC-Sharing}:
	This scheme employs the knowledge graph for clustering.
	In each cluster, a randomly selected personal prompt can be designated as the public prompt.
	\item \emph{\underline{C}lustering-\underline{F}ree \underline{I}nference \underline{S}cheme (CFIS)} \cite{DHY-Sharing2}:
	This scheme treats all UEs as a single cluster, from which a personal prompt is randomly selected as the public one.
	\item \emph{\underline{I}ndependent \underline{I}nference \underline{S}cheme (IIS)} \cite{D2SAC, ZXY-INFOCOM}:
	This scheme does not introduce the common inference phase.
\end{itemize}

\begin{figure}[t]
	\centering
	\includegraphics[width=3in]{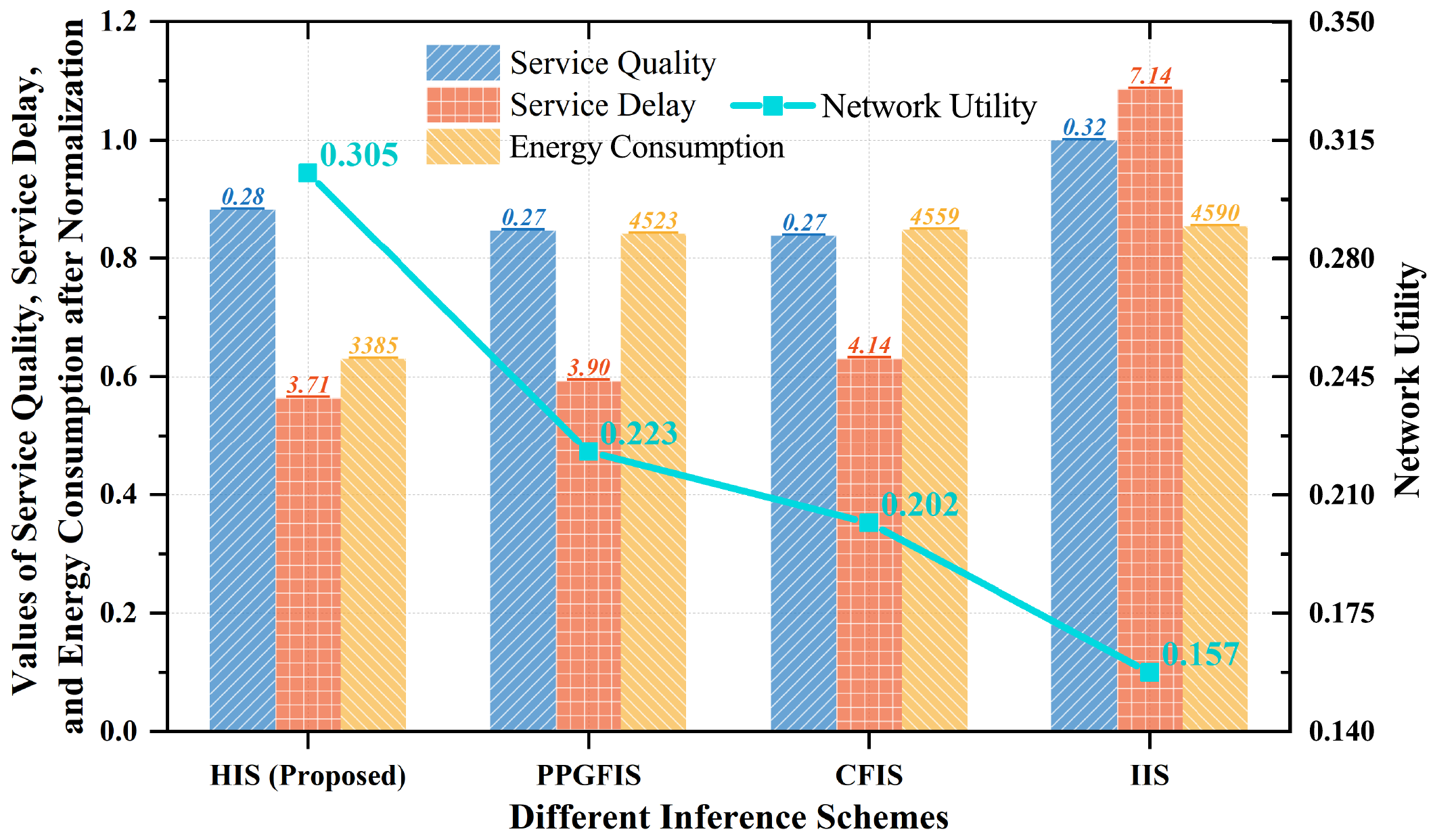}
	\vspace{-3mm}
	\caption{Values of service quality, service delay, energy consumption, and network utility for several inference schemes.}
	\label{fig:11-HIS-Performance} 
\end{figure}

\textit{1) Case Study:}
Sample images generated by three collaborative inference schemes (i.e., HIS, PPGFIS, and CFIS) are shown in Fig. \ref{fig:10-Case-Study}.
\zxxy{Among them, the HIS shows superior performance by maintaining high perceptual quality even with a large number of common inference steps.}
This effectively overcomes a key limitation reported in prior work \cite{XGC-Sharing}, where the number of common inference steps is typically restricted to one-third of the total number of inference steps.

\textit{2) HIS Performance:}
A detailed numerical comparison highlighting the performance of the HIS is presented in \mbox{Fig. \ref{fig:11-HIS-Performance}.}
Compared to the PPGFIS, CFIS, and IIS, the HIS achieves a significantly better trade-off among service quality, service delay, and energy consumption.
These improvements translate into network utility improvements of 36.8\%, 51.0\%, and 94.3\%, respectively.
Notably, the HIS achieves superior service quality while simultaneously reducing both service delay and energy consumption when compared to the PPGFIS and CFIS.
\zxxy{This advantage stems from its efficient utilization of the common inference phase, which allows more common inference steps without degrading the quality of the generated content.}
As a result, the overall inference workload is reduced, enabling faster response times and lower power consumption across the network.

\begin{figure}[t]
	\centering
	\subfigure{
		\includegraphics[width=1.67in, height=1.3in]{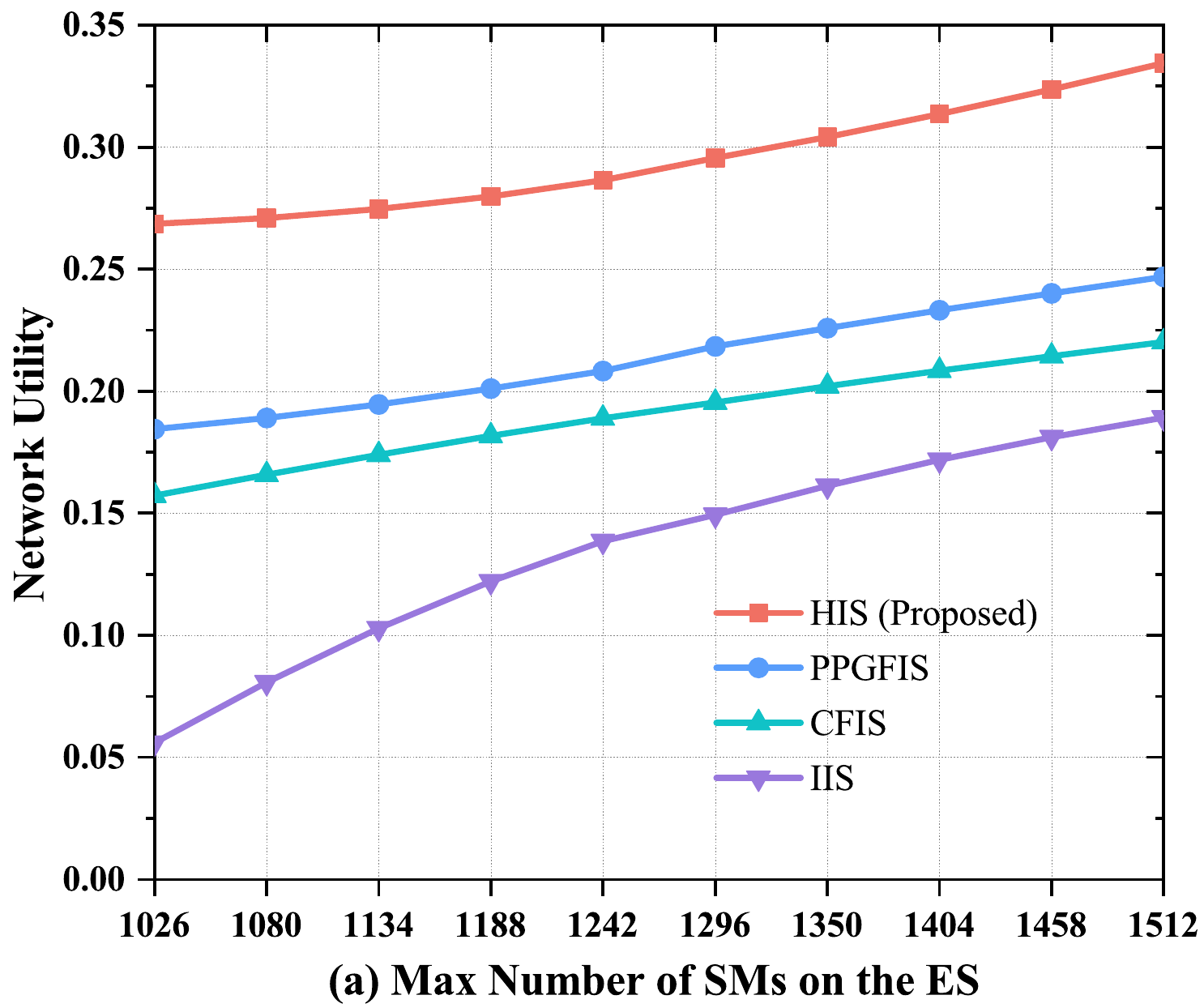}
		\label{fig:12-ES-SM}
	}
	\hspace{-4.3mm}
	\subfigure{
		\includegraphics[width=1.67in, height=1.3in]{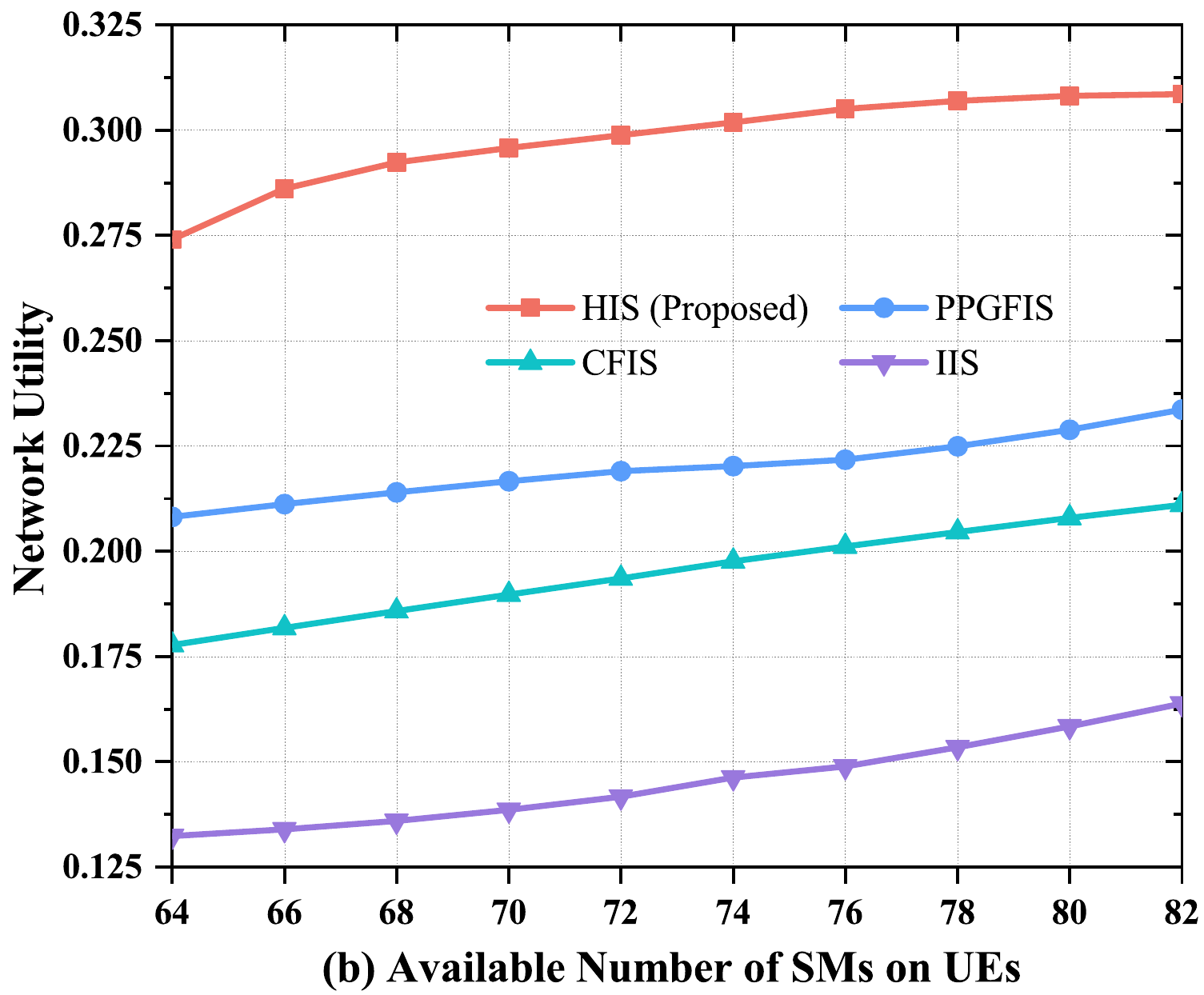}
		\label{fig:12-UE-SM}
	}
	\vspace{-6.5mm}
	\caption{Service robustness of several inference schemes to computational resources. (a) Network utility under varying maximum number of SMs on the ES. (b) Network utility under varying available number of SMs on UEs.}
	\label{fig:12-ES-UE-SM} 
\end{figure}

\textit{3) Service Robustness:}
The service robustness of various inference schemes is evaluated by analyzing their network utility under varying levels of computation resources.
As shown in Fig. \ref{fig:12-ES-UE-SM}, the HIS consistently delivers higher network utility than all benchmark schemes across a broad range of available SMs on both the ES and UEs.
When the computational capacity on the ES fluctuates, the HIS exhibits only a moderate utility degradation of 18.7\%, in contrast to the more substantial drops observed in the PPGFIS (23.1\%), CFIS (29.0\%), and IIS (74.1\%).
A similar trend is observed when UE resources vary: the HIS maintains a low utility reduction of 9.0\%, outperforming the PPGFIS (11.6\%), CFIS (15.3\%), and IIS (14.9\%).
Notably, the HIS also requires fewer computational resources to achieve the same utility level as its counterparts.
These results highlight the strong adaptability of the HIS to dynamic environments, allowing it to sustain high-quality services even under resource constraints.

\begin{figure}[t]
	\centering
	\includegraphics[width=2.92in]{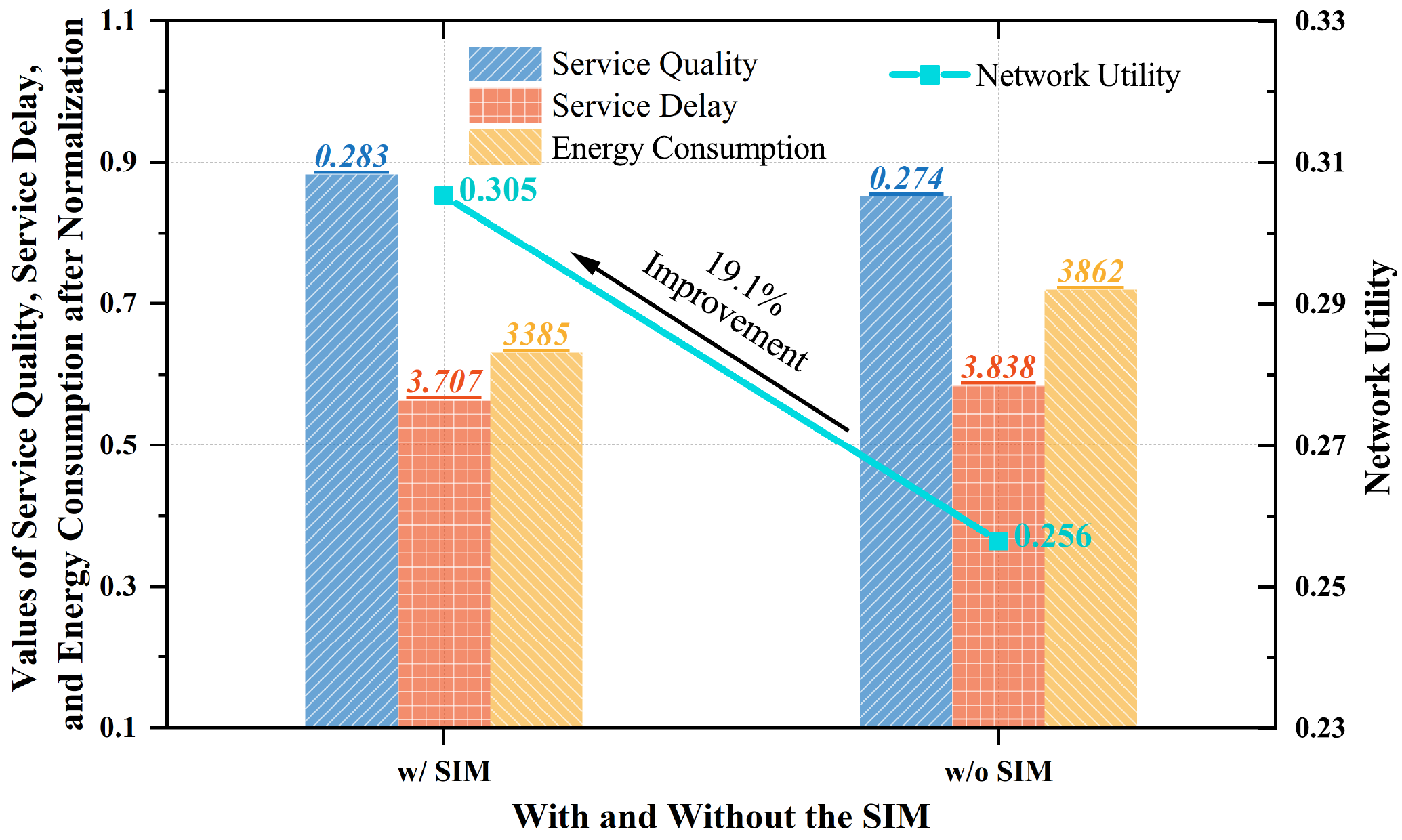}
	\vspace{-3mm}
	\caption{Values of service quality, service delay, energy consumption, and network utility with and without the SIM in the HIS.}
	\label{fig:13-wSIM} 
\end{figure}

\textit{4) Ablation Study:}
Fig. \ref{fig:13-wSIM} illustrates the quantitative impact of the SIM in the HIS.
The empirical results reveal that enabling the SIM leads to a 19.1\% improvement in overall network utility.
This performance gain can be better understood by referring to the quality transition curves shown in Fig. \ref{fig:5-Fitting-Function}.
Specifically, the SIM flattens the quality degradation curve, effectively reducing the sensitivity of service quality to variations in the common inference steps.
As a result, the HIS is able to complete personalized inference with fewer total steps, leading to notable reductions in resource consumption.

\begin{figure}[t]
	\centering
	\includegraphics[width=3in]{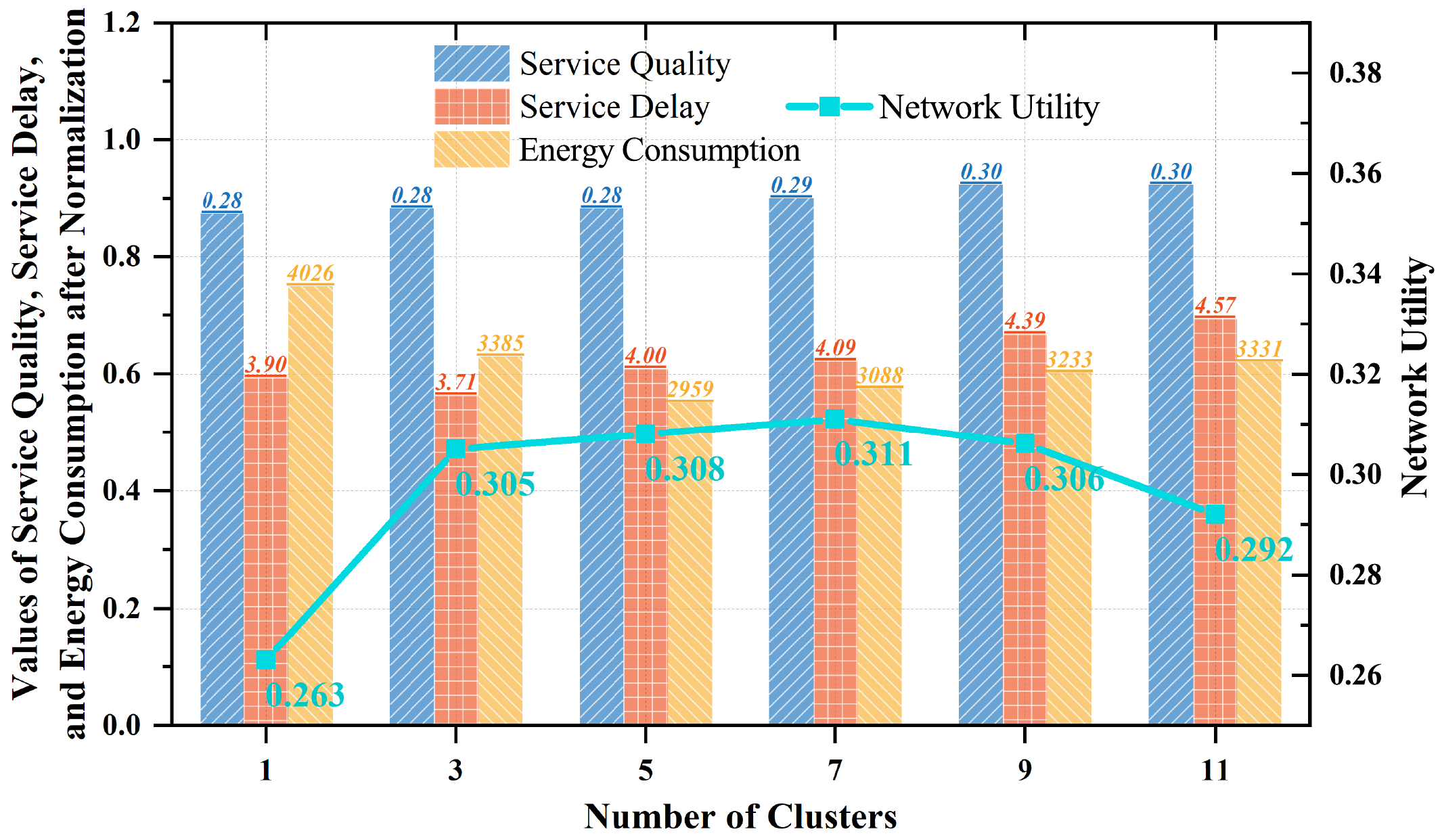}
	\vspace{-3mm}
	\caption{Values of service quality, service delay, energy consumption, and network utility under different numbers of clusters.}
	\label{fig:14-Number-Cluster}
\end{figure}

\textit{5) Network Setting:} Fig. \ref{fig:14-Number-Cluster} presents the impact of the number of clusters.
With fewer clusters, lower public-personal similarity necessitates a smaller number of common inference steps to maintain service quality.
As a result, the total number of executed steps increases, thus exacerbating service delay and energy consumption.
As the number of clusters increases, more similar personal prompts are grouped together, leading to an increase in public-personal similarity.
This allows for a larger number of common inference steps, thereby improving network utility.
However, an excessive number of clusters leads to diminishing returns in public-personal similarity and excessive computational overhead, ultimately degrading network utility.
Furthermore, within a sufficiently broad range, variations in the number of clusters have minimal effect on network utility.
The improvements in public-personal similarity and the corresponding increase in computational overhead largely offset each other.

\subsection{The Performance and Sensitivity of the Algorithms}
In this subsection, we demonstrate the performance and sensitivity of our proposed algorithms.

\begin{figure}[t]
	\centering
	\includegraphics[width=3in]{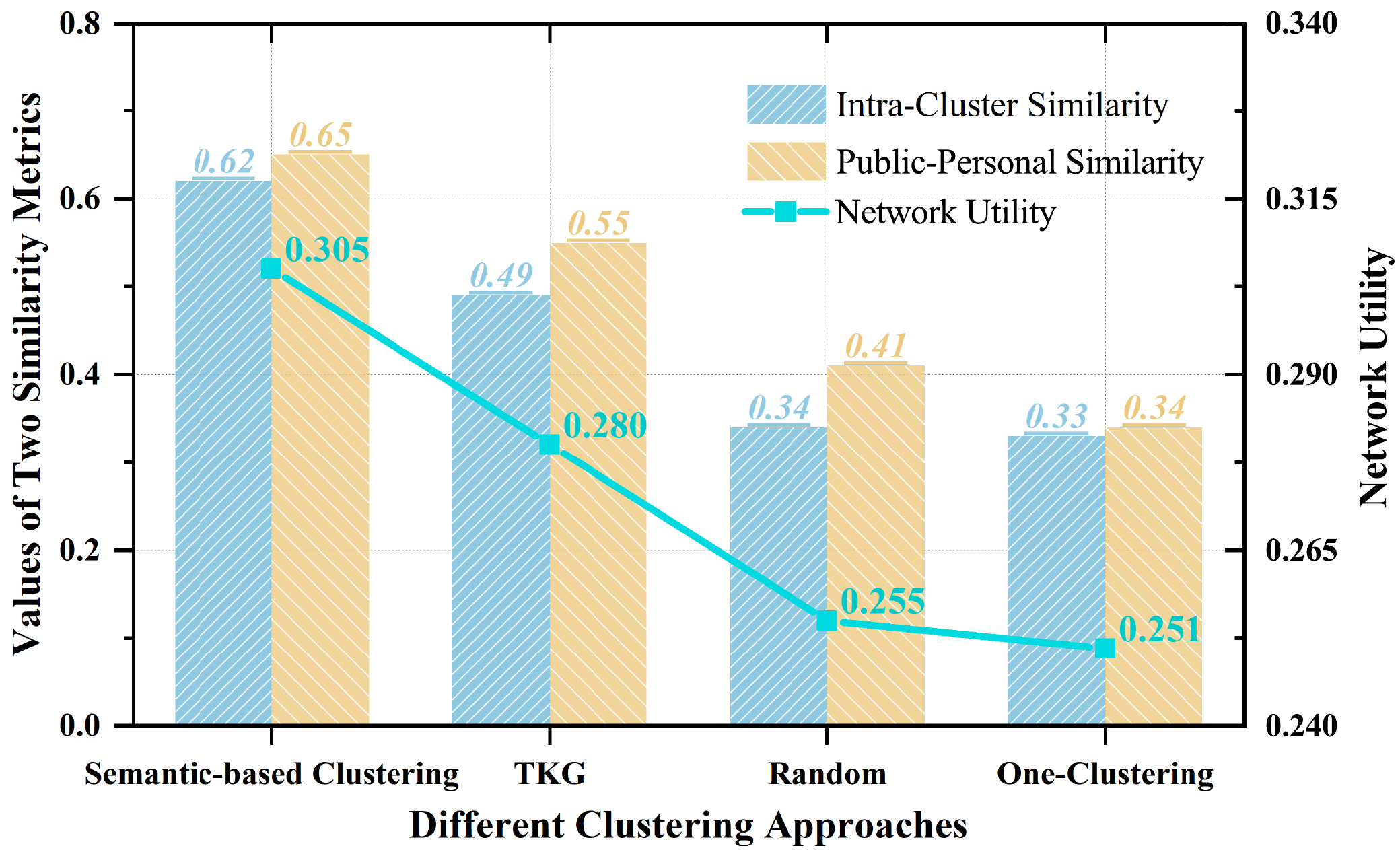}
	\vspace{-3mm}
	\caption{Values of intra-cluster similarity, public-personal similarity, and network utility using different clustering approaches.}
	\label{fig:15-Clustering} 
\end{figure}

\textit{1) Performance of Clustering:}
Fig. \ref{fig:15-Clustering} compares the performance of various clustering approaches using two key evaluation metrics: intra-cluster similarity and public-personal similarity.
Among them, our proposed semantic-based clustering approach demonstrates the best overall performance, achieving a public-personal similarity score of 0.65.
In contrast, the Task Knowledge Graph (TKG)-based approach used in the PPGFIS prioritizes entity-level consistency but fails to account for deeper semantic relationships, resulting in notable degradation in network utility.
Furthermore, baselines such as random clustering and one-clustering lack a principled clustering mechanism.
As a result, the public prompts generated under these approaches fail to represent the semantic diversity of user inputs, severely limiting their effectiveness in guiding the common inference.

\begin{figure}[t]
	\centering
	\includegraphics[width=2.9in]{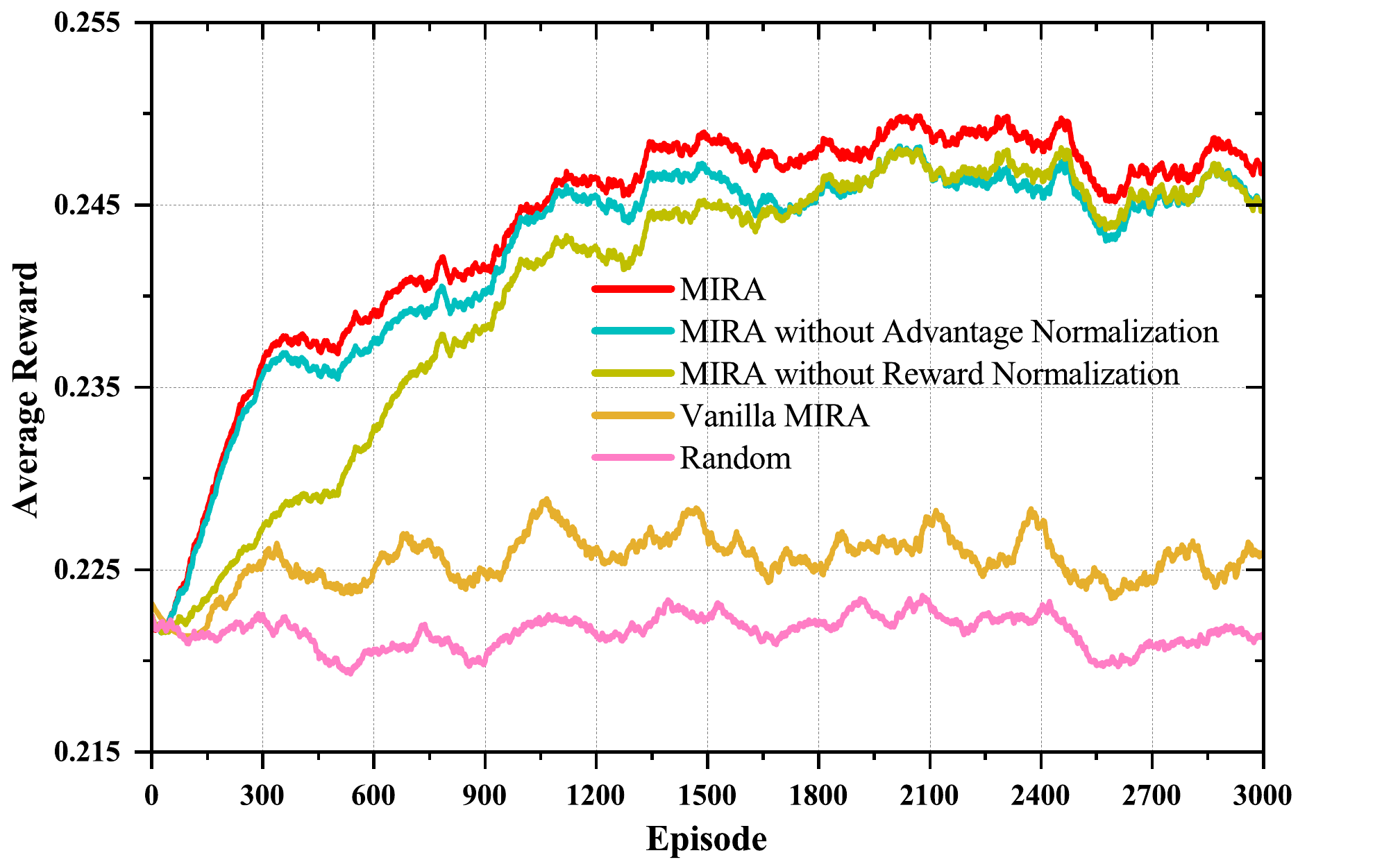}
	\vspace{-3mm}
	\caption{Reward curves of different model inferencing algorithms. All algorithms are based on the optimal resource allocation.}
	\label{fig:16-DRL-Convergence} 
\end{figure}

\textit{2) Performance of Model Inferencing:} Fig. \ref{fig:16-DRL-Convergence} presents the impact of various techniques incorporated into the model inferencing algorithm.
The absence of advantage normalization leads to higher variance in advantage estimates, which amplifies noise and results in degraded performance.
Similarly, without reward normalization, the algorithm struggles to balance reward values, leading to slower convergence.
Vanilla MIRA, which does not incorporate normalization and loss truncation, tends to converge to suboptimal solutions.
This occurs as the lack of these techniques leads the algorithm to overfit noisy gradients, thereby diminishing its ability to generalize effectively and converge to a better solution.

\begin{figure}[t]
	\centering
	\includegraphics[width=3in]{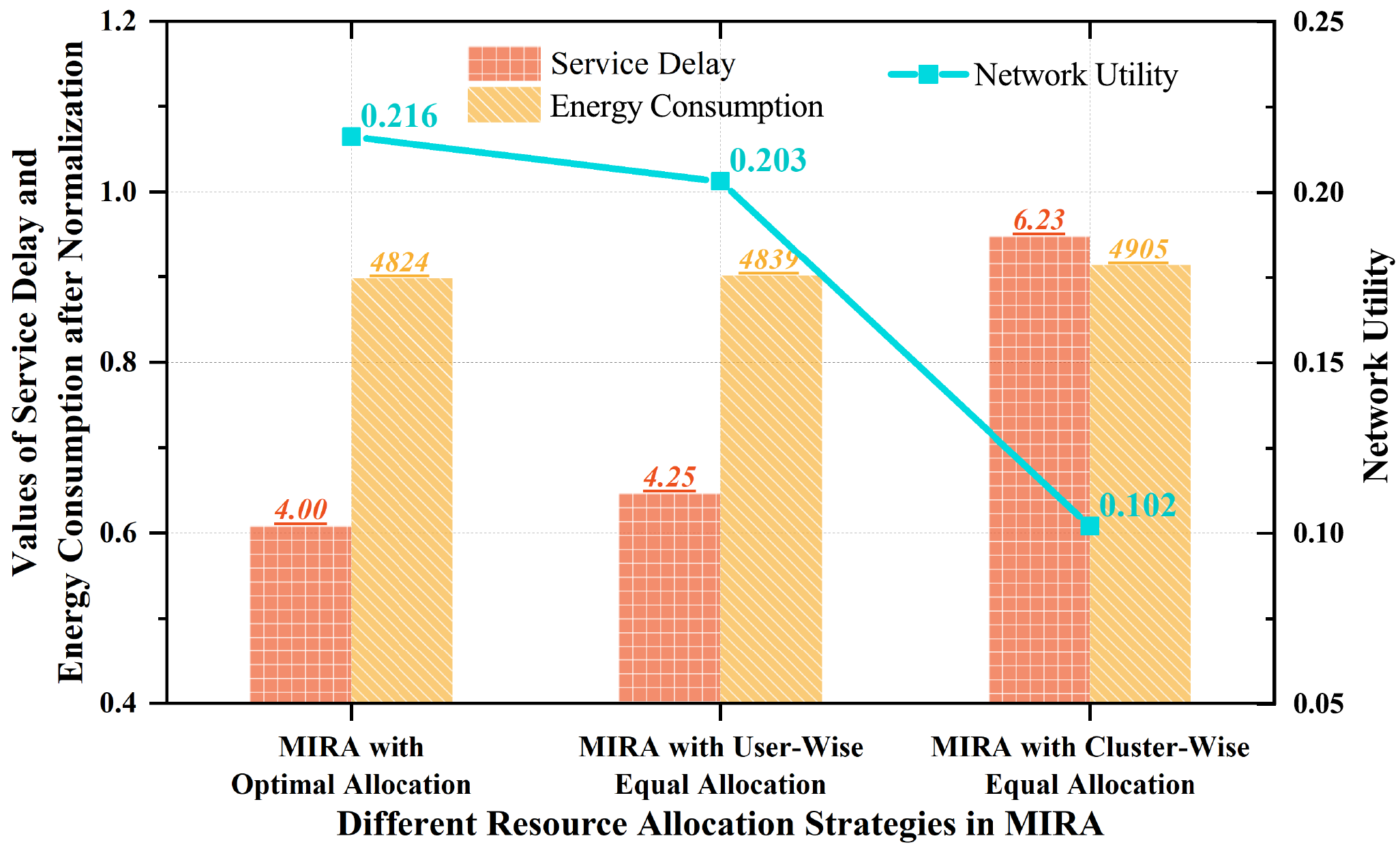}
	\vspace{-3mm}
	\caption{Values of service delay, energy consumption, and network utility using different resource allocation strategies.}
	\label{fig:17-Resource-Allocation} 
\end{figure}

\textit{3) Performance of Resource Allocation:}
We evaluate the effectiveness of our proposed resource allocation strategy in MIRA under scenarios characterized by highly imbalanced user intentions, where the number of UEs varies substantially across clusters.
To understand the performance impact of allocation granularity, we compare our method against two baseline strategies.
In the user-wise equal allocation scheme, the computational resources on the ES are evenly distributed across all UEs.
Conversely, the cluster-wise equal allocation scheme assigns equal resources to each cluster, without accounting for the number of UEs per cluster.
As shown in Fig. \ref{fig:17-Resource-Allocation}, our proposed optimal allocation strategy dynamically adapts to cluster-level demands, significantly improving network performance.
Specifically, it achieves the lowest service delay and energy consumption, leading to the highest network utility value of 0.216.

\begin{figure}[t]
    \centering
 	\includegraphics[width=3in ]{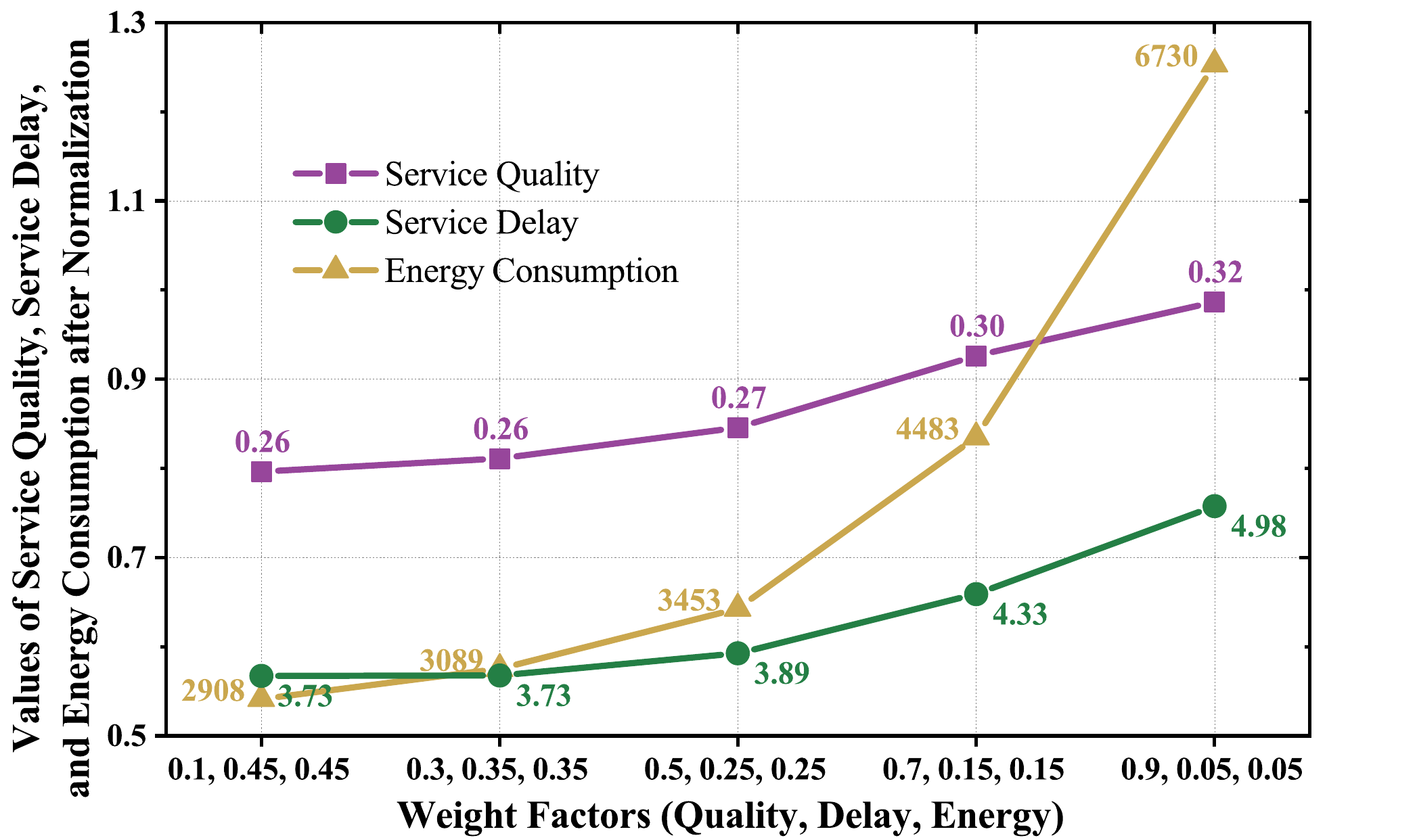}
 	\vspace{-3mm}
 	\caption{Values of service quality, service delay, and energy consumption under different weight factors.}
 	\label{fig:17-Quality-Delay-Energy} 
\end{figure}

\textit{4) Sensitivity to Weight Settings:} \zxxy{As illustrated in Fig. \ref{fig:17-Quality-Delay-Energy}, as the weight factor of service quality increases, the values of service quality, service delay, and energy consumption also rise, indicating that our proposed MIRA algorithm tends to achieve higher service quality with fewer common inference steps.}
This trend shows how the MIRA algorithm adjusts its behavior across different optimization preferences, from quality-oriented to delay- and energy-oriented MEC networks, demonstrating its sensitivity to the weight settings.

\section{Conclusion}\label{sec:Conclusion}

In this work, we proposed an efficient inference scheme, HIS, for MEC-enabled AIGC services. 
In such a scheme, we first introduced an advanced module, SIM, to improve the inference efficiency, and a new metric, HIQM, to quantify the service quality.
Then, we formulated the joint optimization problem of user clustering, model inferencing, and resource allocation.
We further proposed the semantic-based clustering along with the integrated learning and optimization approach for effective problem solving.
Finally, simulation results show that our proposed scheme significantly outperforms existing benchmarks, achieving performance improvements ranging from 36.8\% to 94.3\%.


\begin{IEEEbiography}[{\includegraphics[width=1in,height=1.25in,clip, keepaspectratio]{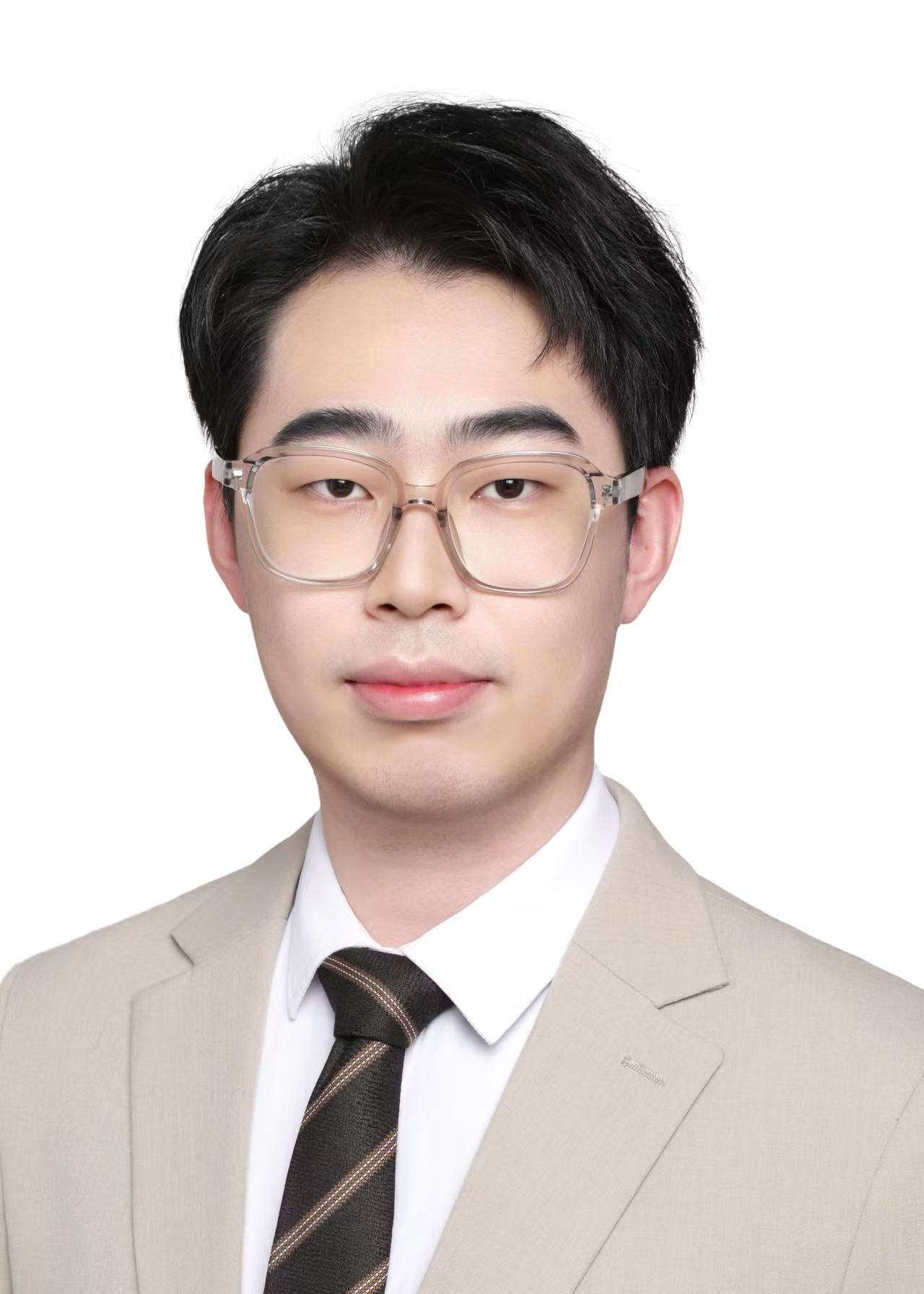}}]{Xinyi Zhuang} (Graduate Student Member, IEEE) is pursuing his Ph.D. degree with the School of Information Science and Technology, Harbin Institute of Technology, Shenzhen, China. He received his B.Eng. degree in Communication Engineering from Northwestern Polytechnical University in 2023. His main research interests include distributed training and inference of large AI models, mobile edge computing, network optimization, and multi-agent reinforcement learning.
\end{IEEEbiography}

\begin{IEEEbiography}[{\includegraphics[width=1in,height=1.25in,clip, keepaspectratio]{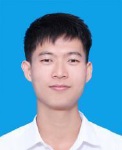}}]{Jiaqi Wu} (Member, IEEE)
is currently a Lecturer with the School of Computer Science, Guangdong University of Finance, Guangzhou, China.
He received the Ph.D. degree in Information and Communication Engineering from Harbin Institute of Technology, Shenzhen, China, in 2025.
His main research interests include network optimization, deep reinforcement learning, edge intelligence, generative AI, and large language model.
\end{IEEEbiography}

\begin{IEEEbiography}[{\includegraphics[width=1in,clip, keepaspectratio]{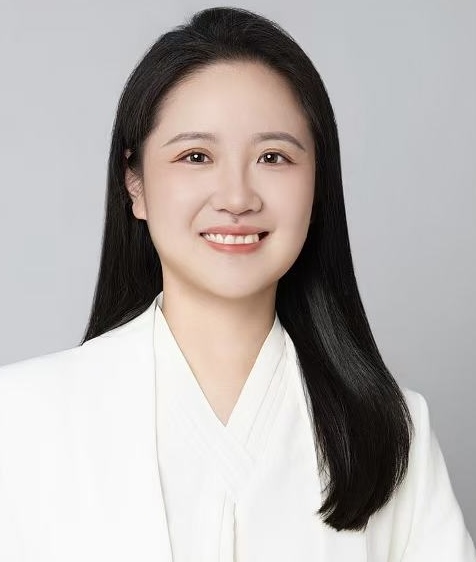}}]{Yuan Luo} (Senior Member, IEEE) received the Ph.D. degree from The Chinese University of Hong Kong, Hong Kong, China, in 2015. She was a visiting scholar with University of California, Berkeley, USA, from 2014 to 2015, a postdoctoral researcher with The Chinese University of Hong Kong from 2015 to 2017, and a research fellow with Imperial College London, from 2017 to 2021. She is currently an assistant professor with the School of Science and Engineering, The Chinese University of Hong Kong, Shenzhen, China. Her research interests include artificial intelligence, mechanism design and low-latency transmission. She was the recipient of the Best Paper Award in the IEEE WiOpt 2014.
\end{IEEEbiography}

\begin{IEEEbiography}[{\includegraphics[width=1in,height=1.25in,clip, keepaspectratio]{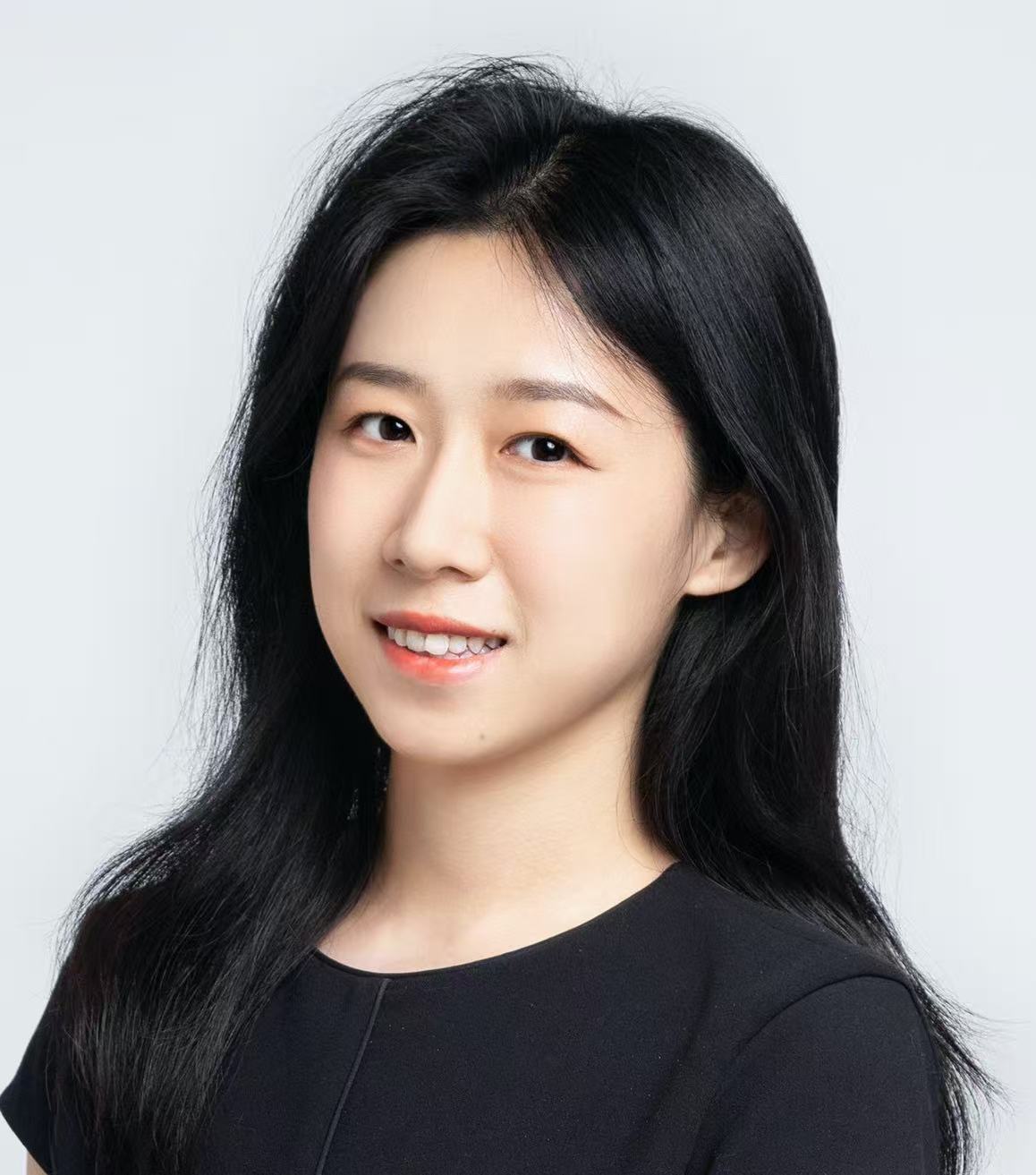}}]{Ming Tang} (Member, IEEE) is an Associate Professor in the Department of Computer Science and Engineering at Southern University of Science and Technology, Shenzhen, China. She received her Ph.D. degree from The Chinese University of Hong Kong, Hong Kong, China, in 2018. She was a postdoctoral research fellow in The University of British Columbia, Vancouver, Canada, from 2018 to 2021. Her research interests include mobile edge computing, edge intelligence, and network economics. She received Best Paper Award from IEEE Transactions on Mobile Computing, IEEE GLOBECOM, and WiOpt. She recently serves as an Editor for IEEE Transactions on Wireless Communications and an Associate Editor for IEEE Internet of Things Journal and IEEE Open Journal of the Communications Society.
\end{IEEEbiography}

\begin{IEEEbiography}[{\includegraphics[width=1in,height=1.25in,clip, keepaspectratio]{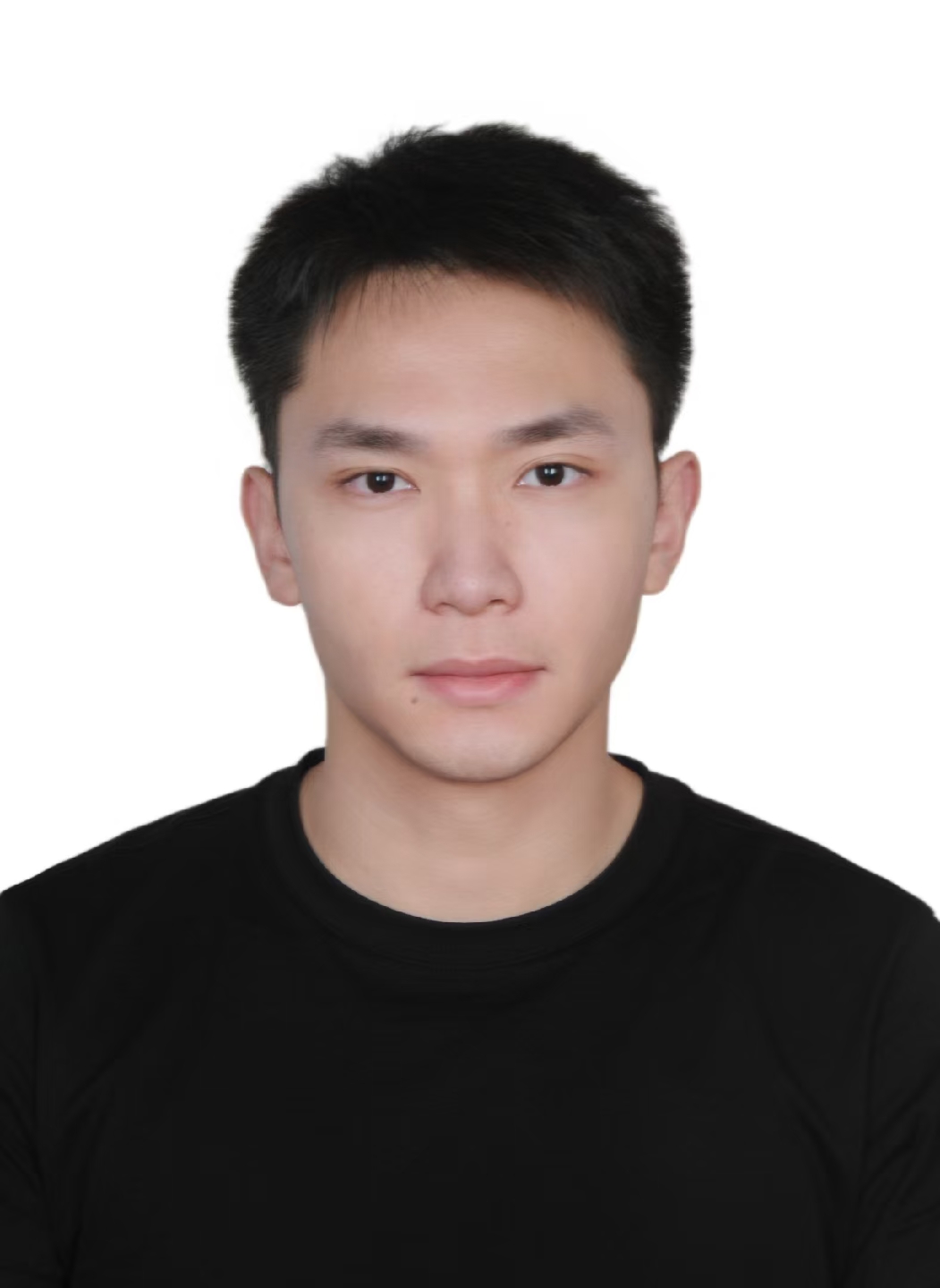}}]{Huaizhe Liu} (Graduate Student Member, IEEE) is currently pursuing the Ph.D. degree with the School of Information Science and Technology, Harbin Institute of Technology, Shenzhen, China.
He received the M.S. degree in Information and Communication Engineering from Harbin Institute of Technology, Shenzhen, China, in 2020.
His main research interests are in the interdisciplinary area between stochastic network optimization, algorithmic game theory, and artificial intelligence.
\end{IEEEbiography}

\begin{IEEEbiography}[{\includegraphics[width=1in,height=1.25in,clip, keepaspectratio]{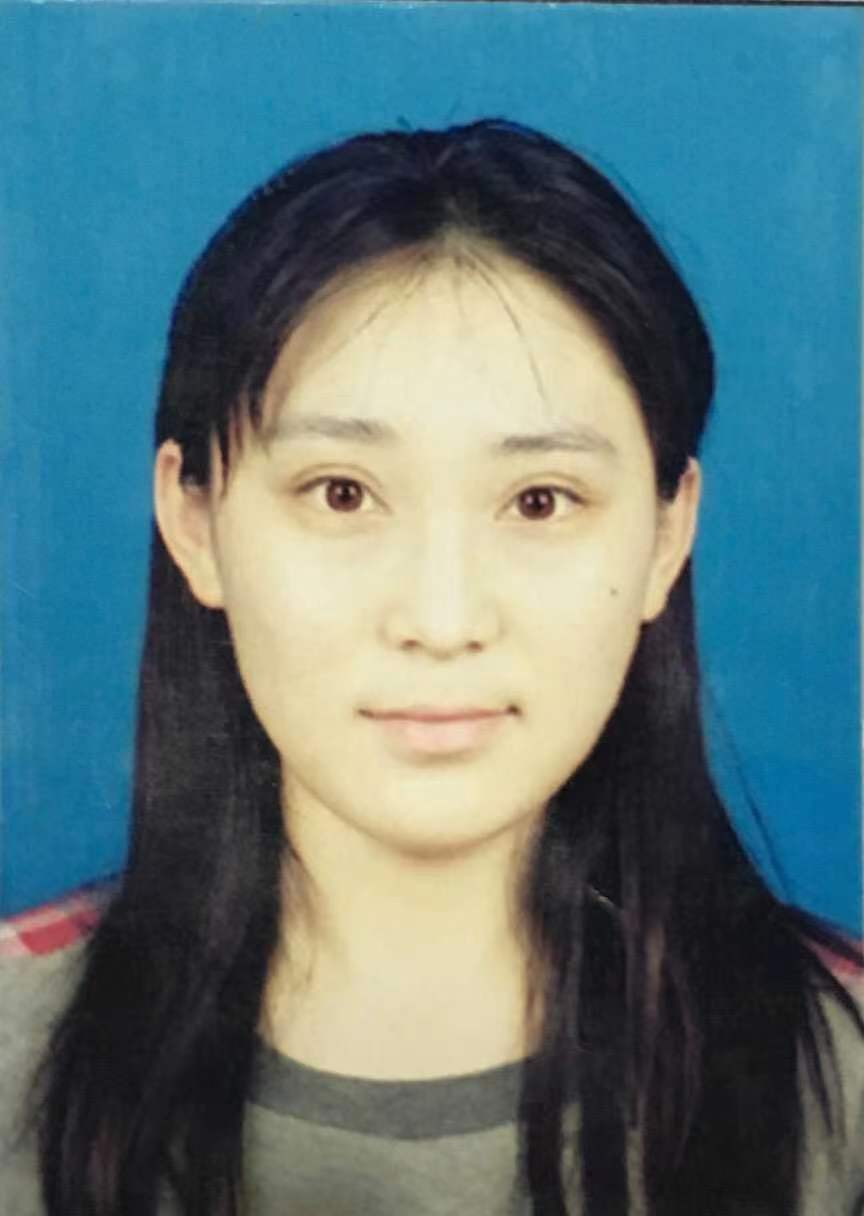}}]{Hongjia Wu} (Member, IEEE) received the Ph.D. degree in computer
science from the National University of Defense
Technology, China, in 2023. She was a visiting student with the Singapore University of Technology and
Design. She is currently a post-doctoral fellow with
the Education University of Hong Kong in China.
Her research interests include edge intelligence, game theory, and the Internet of Things (IoT).
\end{IEEEbiography}

\begin{IEEEbiography}[{\includegraphics[width=1in,height=1.25in,clip, keepaspectratio]{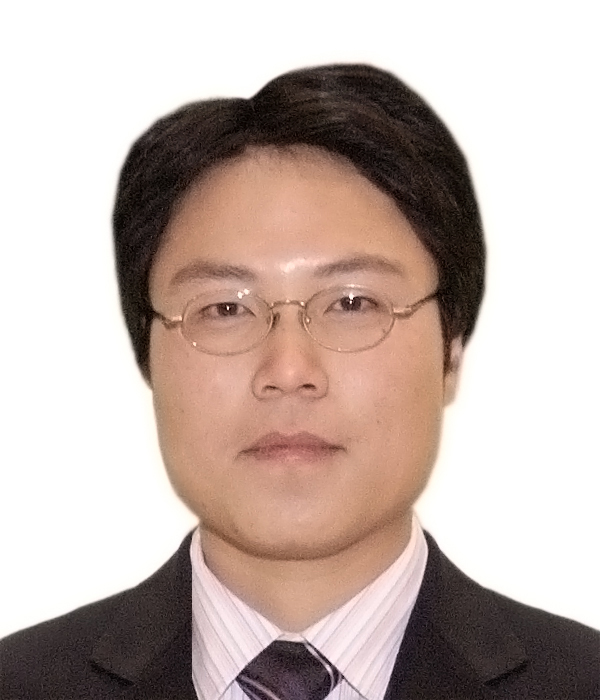}}]{Lin Gao} (Senior Member, IEEE) is a Professor with the School of Information Science and Technology, Harbin Institute of Technology, Shenzhen, China. He received the Ph.D. degree in Electronic Engineering from Shanghai Jiao Tong University, Shanghai, China, in 2010. His main research interests are in the interdisciplinary area between optimization, game theory, and artificial intelligence, with particular focuses on reinforcement learning, federated learning, crowd/edge intelligence, mobile edge computing, and cognitive networking. He is the co-recipient of 6 Best Paper Awards from leading conference proceedings on wireless communications and networking, including IEEE GLOBECOM 2026, WiOpt 2025, and WCNC 2019. He received the IEEE ComSoc Asia-Pacific Outstanding Young Researcher Award in 2016.
\end{IEEEbiography}

\begin{IEEEbiography}[{\includegraphics[width=1in,height=1.25in,clip, keepaspectratio]{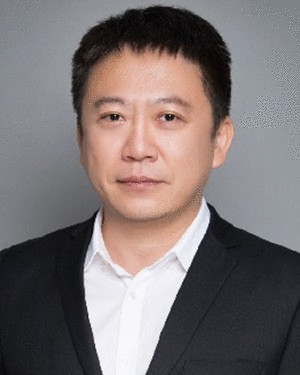}}]{Qinyu Zhang} (Senior Member, IEEE) received
the bachelor's degree in communication engineering from Harbin Institute of Technology, Harbin,
China, in 1994, and the Ph.D. degree in biomedical and electrical engineering from the University
of Tokushima, Tokushima, Japan, in 2003. From
1999 to 2003, he was an Assistant Professor with
the University of Tokushima. Since 2003, he has
been with Harbin Institute of Technology-Shenzhen
(HITsz), Shenzhen, China, where he is currently a
Full Professor and the Vice President of HITsz. He
is also a Professor with the Pengcheng Laboratory, Shenzhen, and the Director
of Guangdong Provincial Key Laboratory of Aerospace Communication and
Networking Technology, Shenzhen. His research interests include aerospace
communications and networks, and wireless communications and networks.
He was awarded the National Science Fund for Distinguished Young Scholars,
the Young and Middle-aged Leading Scientist of China, Chinese New Century
Excellent Talents in University, and three scientific and technological awards
from the provincial/ministerial governments.
\end{IEEEbiography}

\newpage

\section{Appendix}

\subsection{Proof of Theorem 1}\label{sec:Appendix}
The Lagrange function of sub-problem $\mathbb{P}_{\text{II}}^{\text{B}}$ can be defined as:
\begin{equation}
	\mathcal{L}({\bf{V}}, \beta) = \sum_{m=1}^M \left( \frac{\alpha_m^{(1)} + \alpha_m^{(2)}}{v_m^{\text{ES}}} \right)
	+ \beta \left( \sum_{m=1}^M v_{m}^{\text{ES}} - v_{\max}^{\text{ES}} \right),
\end{equation}
where $\beta$ is the Lagrange multiplier.

The Karush-Kuhn-Tucker (KKT) conditions are as follows:

$\cdot \ \ \text{Stationarity}$:
\begin{equation} \label{eq:Stationarity}
	\frac{\partial \mathcal{L}}{\partial v_m^{\text{ES}}} = -\frac{\alpha_m^{(1)} + \alpha_m^{(2)}}{(v_m^{\text{ES}})^2} + \beta = 0, \quad \forall m \in \mathcal{M}.
\end{equation}

$\cdot \ \ \text{Complementary\ Slackness}$:
\begin{equation} \label{eq:Complementary-Slackness}
	\beta \left( \sum_{m=1}^M v_{m}^{\text{ES}} - v_{\max}^{\text{ES}} \right) = 0.
\end{equation}

$\cdot \ \ \text{Primal\ Feasibility}$:
\begin{equation}\label{eq:Primal-Feasibility}
	\sum_{m=1}^M v_m^{\text{ES}} \leq v_{\max}^{\text{ES}}.
\end{equation}

$\cdot \ \ \text{Dual\ Feasibility}$: $\beta \geq 0$.

According to Eq. \eqref{eq:Stationarity}, we have:
\begin{equation}\label{eq:v_m}
	v_m^{\text{ES}} = \sqrt{\frac{\alpha_m^{(1)} + \alpha_m^{(2)}}{\beta}}, \quad \forall m \in \mathcal{M}.
\end{equation}

Since the objective function is strictly decreasing with respect to each positive $v_m^{\text{ES}}$, the resource constraint is active at the optimum, i.e., $\sum_{m=1}^M v_m^{\text{ES}} = v_{\max}^{\text{ES}}$.
Considering Eq. \eqref{eq:Primal-Feasibility}, we have:
\begin{equation}\label{eq:beta}
	\beta = \left( \frac{\sum_{m=1}^M \sqrt{\alpha_m^{(1)} + \alpha_m^{(2)}}}{v_{\max}^{\text{ES}}} \right)^2.
\end{equation}

Based on Eq. \eqref{eq:v_m} and Eq. \eqref{eq:beta}, we have:
\begin{equation}
	(v_m^{\text{ES}})^{\star} = v_{\max}^{\text{ES}} \frac{\sqrt{\alpha_m^{(1)} + \alpha_m^{(2)}}}{\sum_{m \in \mathcal{M}} \sqrt{\alpha_m^{(1)} + \alpha_m^{(2)}}} ,\quad \forall m \in \mathcal{M}.
\end{equation}

\end{document}